\documentclass[sigconf]{acmart}
\usepackage{natbib}
\AtBeginDocument{%
  }

\copyrightyear{2026}
\acmYear{2026}
\setcopyright{cc}
\setcctype{by}
\acmConference[KDD '26]{Proceedings of the 32nd ACM SIGKDD Conference on
  Knowledge Discovery and Data Mining V.2}{August 09--13, 2026}{Jeju Island,
  Republic of Korea}
\acmBooktitle{Proceedings of the 32nd ACM SIGKDD Conference on Knowledge
  Discovery and Data Mining V.2 (KDD '26), August 09--13, 2026, Jeju Island,
  Republic of Korea}
\acmDOI{10.1145/3770855.3818956}
\acmISBN{979-8-4007-2259-2/2026/08}
\begin{document}

\title{From Causal Discovery to Dynamic Causal Inference in Neural Time Series}

\author{Dmitry Zaytsev}
\orcid{0000-0002-0902-3896}
\affiliation{%
  \institution{Lucy Family Institute for Data \& Society \\ University of Notre Dame}
  \city{Notre Dame}
  \state{Indiana}
  \country{USA}
}
\email{zaytsevdi2@gmail.com}

\author{Valentina V. Kuskova}
\orcid{0000-0003-4716-2544}
\affiliation{%
 \institution{Lucy Family Institute for Data \& Society \\ University of Notre Dame}
 \city{Notre Dame}
 \state{Indiana}
 \country{USA}}
\email{vkuskova@nd.edu}

\author{Michael Coppedge}
\orcid{0000-0002-4865-6901}
\affiliation{%
 \institution{Political Science \\ University of Notre Dame}
 \city{Notre Dame}
 \state{Indiana}
 \country{USA}}
\email{mcoppedg@nd.edu}

\renewcommand{\shortauthors}{Dmitry Zaytsev, Valentina V. Kuskova, \& Michael Coppedge}

\begin{abstract}
Time-varying causal models provide a powerful framework for studying dynamic scientific systems, yet most existing approaches assume that the underlying causal network is known a priori - an assumption rarely satisfied in real-world domains where causal structure is uncertain, evolving, or only indirectly observable. This limits the applicability of dynamic causal inference in many scientific settings.
We propose Dynamic Causal Network Autoregression (DCNAR), a two-stage neural causal modeling framework that integrates data-driven causal discovery with time-varying causal inference. In the first stage, a neural autoregressive causal discovery model learns a sparse directed causal network from multivariate time series. In the second stage, this learned structure is used as a structural prior for a time-varying neural network autoregression, enabling dynamic estimation of causal influence without requiring pre-specified network structure.
We evaluate the scientific validity of DCNAR using behavioral diagnostics that assess causal necessity, temporal stability, and sensitivity to structural change, rather than predictive accuracy alone. Experiments on multi-country panel time-series data demonstrate that learned causal networks yield more stable and behaviorally meaningful dynamic causal inferences than coefficient-based or structure-free alternatives, even when forecasting performance is comparable. These results position DCNAR as a general framework for using AI as a scientific instrument for dynamic causal reasoning under structural uncertainty.
\end{abstract}





\maketitle

\section{Introduction}

Scientific models are not evaluated solely by how accurately they predict future observations, but by what they reveal about the underlying processes that generate data. This distinction is especially important in the social sciences and economics, where researchers seek to model human systems as complex adaptive processes and to evaluate the dynamic consequences of policy-relevant interventions through counterfactual simulation rather than prediction alone. In such settings, predictive accuracy is a necessary condition for model credibility, but it is not sufficient for scientific explanation.

Dynamic causal understanding therefore relies on forms of model interrogation that go beyond forecast performance \cite{Buehner12}. Scientists examine impulse response functions \citep{Lutkepohl18,Potter00} to study how shocks propagate over time, counterfactual trajectories \citep{Weber08, Morgan14} to assess the effects of alternative interventions, and the stability and persistence of causal effects across regimes \cite{Cirone22}. These tools support explanation, hypothesis testing, and early warning in ways that aggregate error metrics cannot capture.

Despite this, much of modern machine learning for time series prioritizes predictive performance \cite{Masini23}. Flexible neural models can achieve strong forecasting results \cite{Faheem24} but often behave poorly under causal interrogation \cite{Karaca22}. Impulse responses may exhibit oscillations or sign reversals \citep{Nounou06, Inoue13}, counterfactual trajectories may diverge unrealistically \cite{Varshney}, and inferred dynamics can be highly sensitive to small perturbations \cite{Ivanovs}. Consequently, such models are difficult to use as scientific instruments, even when their predictive accuracy is competitive.

Existing frameworks for dynamic causal inference face a complementary limitation. Models that provide interpretable time-varying causal parameters, such as time-varying vector autoregressions or network autoregressive models, typically assume that the underlying causal structure is known in advance \cite{Ding25}. In many domains, including political systems \cite{Blackwell13}, ecological networks \cite{Damos24}, and social processes \cite{Ross24}, this assumption is rarely satisfied. Causal structure is often uncertain, evolving, or itself the object of investigation. Researchers are therefore forced to choose between interpretable dynamic models with questionable structural assumptions and flexible predictive models that offer little causal insight \citep{Smart25, Rudin22, Engel21}.

This paper introduces \emph{Dynamic Causal Network Autoregression} (DCNAR), a framework designed to bridge this divide. DCNAR enables dynamic causal analysis when causal structure is unknown by integrating data-driven causal discovery with time-varying network autoregression in a principled pipeline. Rather than optimizing exclusively for predictive accuracy, DCNAR is designed to support scientifically meaningful interrogation through impulse responses and counterfactual analysis, accepting modest trade-offs in predictive optimality in exchange for greater interpretability, stability, and theoretical coherence of inferred causal dynamics.

Our methodology is motivated by domain-specific democracy panel data characterized by many short, heterogeneous country-level time series, whose constraints directly inform the design of DCNAR and its emphasis on structural regularization. Using multi-country panel time-series data as a motivating scientific application, we show that DCNAR produces impulse responses and counterfactual trajectories that are stable, interpretable, and consistent with theoretical expectations, even when causal structure is not known a priori. While DCNAR does not dominate all baselines on every predictive metric, it remains competitively calibrated and avoids the pathological causal behavior often exhibited by purely predictive models. These properties make DCNAR suitable not merely as a forecasting tool, but as a methodological instrument for scientific reasoning about dynamic systems.

The contributions of this work are twofold. First, we demonstrate that DCNAR achieves competitive predictive and distributional performance relative to common time-series baselines while maintaining probabilistic calibration. Second, and more importantly, we show that DCNAR uniquely supports interpretable, stable, and theoretically meaningful impulse-response and counterfactual analysis in settings with unknown causal structure, enabling forms of dynamic causal understanding that are inaccessible to existing approaches.

\section{Scientific Challenge: Dynamic Causality Without Known Structure}
\label{sec:challenge}

Many scientific domains seek to understand not only whether variables are related, but how causal relationships evolve over time \cite{Thagard12}. In complex systems \cite{Wagner99} such as political regimes or social processes, causal influence is rarely static: interactions may strengthen, weaken, or reverse as contextual conditions change \cite{Andersson19}. Capturing such temporal variation is therefore essential for scientific explanation, early warning, and counterfactual reasoning \citep{Engel21, Hoffman21}. Time-varying causal models, including time-varying vector autoregression \cite{Haslbeck21} and network autoregression \citep{Zhu17, Armillotta23}, have been widely developed for this purpose \cite{Clare19}.

Despite their conceptual appeal, the applicability of dynamic causal models is constrained by a central assumption: the underlying causal network is known in advance \citep{Ding25, Defilippo24, Rottman14}. Most formulations require researchers to specify which directed interactions are admissible based on theory, expert knowledge, or prior empirical studies. In many scientific domains, however, causal structure is precisely what is uncertain, contested, or under investigation \citep{Gerring05, Kurki08}. As a result, researchers face a persistent trade-off between imposing potentially incorrect structural assumptions or abandoning explicit causal modeling in favor of flexible predictive approaches that sacrifice interpretability and hypothesis testing \citep{Cranmer17, Ackley22}.

At a deeper level, this trade-off reflects a structural rather than algorithmic limitation. Joint estimation of time-varying causal effects and network structure is ill-posed in finite, high-dimensional time series, while unconstrained dynamic models are unstable and difficult to interpret. Existing approaches therefore stabilize estimation by fixing structure \cite{Bongers18}, typically treating causal discovery and dynamic inference as separate problems \cite{Glymour19}. This separation transforms dynamic causal models into descriptive tools conditioned on assumed networks \cite{Friston19}, limiting their use for evaluating or discovering causal hypotheses. The challenge addressed in this work is to overcome this limitation by treating causal structure not as a fixed prerequisite, but as a learned and empirically testable component of dynamic causal inference.

\section{Scientific Evidence in Dynamic Causal Models}
\label{sec:sciev}

Dynamic causal models are often evaluated using predictive criteria such as forecast error or likelihood \cite{Rossi24}, but these metrics alone are insufficient for scientific inference. In practice, researchers assess causal models by how they behave under perturbation, examining impulse responses, counterfactual trajectories \citep{Herd19, Benhamza25}, function-valued influence \cite{Kuskova26b}, and the stability of inferred relationships across contexts. Such behavioral probes reveal how effects propagate through a system over time and provide forms of evidence essential for explanation and hypothesis evaluation that cannot be reduced to predictive performance.

\subsection{Limits of prediction-centric evaluation}

Prediction-focused evaluation treats the data-generating process as a black box \cite{Molnar24}, emphasizing forecast accuracy \cite{Morlidge13} without regard to the mechanisms through which predictions are achieved. For causal analysis, this criterion is too weak \cite{Caporale97}. Models with similar predictive accuracy may encode fundamentally different assumptions about feedback, temporal dependence, and causal pathways, leading to divergent conclusions when used for explanation or intervention analysis \citep{Cranmer17, Hofman17}.

These limitations are especially pronounced for flexible machine learning models. Highly expressive architectures can interpolate observed data effectively \cite{Sedler23}, yet exhibit unstable or implausible behavior under causal interrogation \cite{Rawal25}. Small perturbations may induce oscillatory impulse responses, sign reversals, or explosive counterfactual trajectories that are difficult to reconcile with domain knowledge \citep{Tan22, Ruge25}. Consequently, predictive accuracy should be viewed as a necessary but not sufficient condition for scientific usefulness in dynamic causal modeling. Additional criteria are required to assess whether inferred dynamics are coherent, interpretable, and robust.

Moreover, most evaluations of such models have been conducted in technical or engineered domains with controlled dynamics and abundant data \cite{Tan22}. In the social sciences, where systems are adaptive, data are observational, and causal structure is uncertain, systematic evidence on how flexible predictive models behave under causal interrogation remains limited \cite{Ruge25}.

\subsection{Impulse responses and counterfactual behavior}

Impulse response functions are a foundational tool for dynamic causal analysis \cite{Swanson97}. They characterize how localized shocks propagate through a system over time, revealing both the direction and persistence of causal influence. Importantly, impulse responses expose temporal structure that is not visible in static summaries or aggregate performance metrics.

From a scientific perspective, interpretable impulse responses exhibit recognizable qualitative properties \cite{Inoue13}: effects evolve smoothly, decay or persist in ways consistent with known mechanisms, and maintain sign consistency unless substantive reversals are expected \cite{Montoya24}. When impulse responses display erratic oscillations, abrupt sign changes, or sensitivity to minor perturbations, their scientific value is limited \cite{Montoya24}. Evaluating impulse response behavior therefore provides a direct test of whether a model encodes plausible mechanisms rather than merely fitting observed trajectories.

Closely related are counterfactual trajectories, which trace system evolution under hypothetical interventions or shocks \cite{Herd19}. Whereas impulse responses focus on marginal effects, counterfactual analysis captures how perturbations propagate through the full system, potentially affecting many variables simultaneously \cite{Prosperi20}. For scientific reasoning, counterfactual trajectories should remain stable, bounded, and interpretable over time \cite{Weilbach24}, reflecting whether effects amplify, dissipate, or reconfigure \cite{Wickens01}. Counterfactual paths that exhibit implausible volatility or collapse indicate unreliable causal structure \cite{Gerstenberg21}. Because counterfactual behavior is not directly constrained by observed outcomes, it constitutes a stringent test of a model’s internal coherence and causal plausibility.

\subsection{Stability as causal evidence}

A further dimension of scientific evidence concerns stability \cite{Thornes20}. Causal conclusions should not hinge on idiosyncratic features of a particular sample, nor change dramatically under modest perturbations of the data or model specification. In dynamic settings, instability may appear as large fluctuations in inferred effects across time, regimes, or small structural variations.

Stability is therefore central to causal credibility \cite{Bergh24}. Models that produce qualitatively consistent impulse responses and counterfactual trajectories across reasonable perturbations provide stronger evidence of underlying mechanisms than models whose behavior is highly sensitive. This notion of stability is behavioral rather than parametric and is distinct from classical statistical significance, which concerns sampling variability rather than robustness of dynamic behavior \cite{Halpern16}.

\subsection{Implications for model evaluation}

All these considerations suggest that evaluating dynamic causal models requires a broader evidentiary framework than prediction alone. Scientifically useful models should support coherent impulse responses, interpretable counterfactual trajectories, and stable behavior under perturbation, while maintaining reasonable predictive performance \cite{James24}. No single metric captures these properties; instead, they must be assessed through targeted behavioral diagnostics that probe how models respond to interventions and structural variation.

This perspective motivates the framework introduced in the following sections. Rather than treating prediction accuracy as the primary objective, we evaluate dynamic causal models based on the quality and stability of the causal behaviors they induce, allowing models to be assessed as instruments for scientific understanding rather than solely as forecasting tools.

\section{DCNAR: Learned Networks as Structural Priors for Dynamic Causal Modeling}
\label{sec:dcnar}

\subsection{Dynamic causality with unknown structure}

Let
\begin{equation}
\mathbf{x}_t = \left(x_{1,t}, \dots, x_{N,t}\right)^\top, \quad t = 1,\dots,T,
\end{equation}
denote an $N$-dimensional multivariate time series \cite{Ekstrom81}.
Dynamic causal modeling aims to characterize how directed dependencies among variables evolve over time, rather than assuming static interactions.

A general time-varying autoregressive representation \cite{Bringmann17} can be written as
\begin{equation}
\mathbf{x}_t = \sum_{\ell=1}^{L} \mathbf{A}_{\ell}(t)\,\mathbf{x}_{t-\ell} + \boldsymbol{\varepsilon}_t,
\label{eq:tvar_general}
\end{equation}
where $\mathbf{A}_{\ell}(t) \in \mathbb{R}^{N\times N}$ encodes directed causal influence at lag $\ell$ and time $t$.
Without additional constraints, estimating~\eqref{eq:tvar_general} is ill-posed in finite samples, particularly in short, heterogeneous, or nonstationary systems \cite{Horsky17}.

Most existing dynamic causal models address this by assuming that the support of $\mathbf{A}_{\ell}(t)$ is known in advance via a fixed adjacency matrix \cite{Friston03}
\begin{equation}
\mathbf{G} \in \{0,1\}^{N\times N},
\end{equation}
such that
\begin{equation}
A_{\ell,ij}(t) = 0 \quad \text{whenever} \quad G_{ij} = 0.
\label{eq:known_structure}
\end{equation}
This assumption presumes that the causal network is known, fixed, and correct \cite{Rottman14}. In many scientific domains, however, causal structure is uncertain and itself a target of inquiry. DCNAR addresses this gap by replacing assumed structure with a learned and testable structural prior.

\subsection{Overview of the DCNAR framework}

\emph{Dynamic Causal Network Autoregression (DCNAR)} is a two-stage framework that decouples causal structure discovery from  dynamic causal estimation. Rather than jointly estimating a dense time-varying interaction tensor, DCNAR first infers a sparse directed network from data and then uses this network to constrain dynamic inference. This design allows causal influence to vary over time while maintaining interpretability and stability, and it enables impulse-response and counterfactual analysis under structural uncertainty.

\subsection{Stage I: Neural autoregressive causal network discovery}

In the first stage, DCNAR infers a candidate causal network using a neural additive autoregressive causal discovery model \cite{Bussmann21}.
For each target variable $x_{i,t}$, the model takes the form
\begin{equation}
x_{i,t} =
\sum_{j=1}^{N}\sum_{\ell=1}^{L}
f_{ij\ell}\!\left(x_{j,t-\ell}\right) + \varepsilon_{i,t},
\label{eq:navar}
\end{equation}
where $f_{ij\ell}(\cdot)$ are univariate neural functions. The additive structure ensures that contributions from individual lagged variables are separable and interpretable.

Sparsity is induced through regularization on the collection $\{f_{ij\ell}\}$.
Directed influence from variable $j$ to $i$ is summarized via an aggregated causal score
\begin{equation}
S_{ij} = \sum_{\ell=1}^{L} \left\lVert f_{ij\ell} \right\rVert,
\label{eq:score}
\end{equation}
yielding a matrix $\mathbf{S} \in \mathbb{R}^{N\times N}$ of directed causal scores.

The causal score matrix does not represent regression coefficients or estimands with known sampling distributions. It is a sparse, directed Granger-causal network learned from multivariate time series, capturing predictive dependencies across variables rather than identified structural causal effects \cite{Bussmann21}. In DCNAR, it is treated as a \emph{structural hypothesis} rather than as a final inferential object. 

In practice, additional processing steps may be applied to $\mathbf{S}$, including stability assessment and necessity-based screening, to construct a sparse adjacency matrix. In the experiments reported here, we use a weighted adjacency matrix obtained via edge ablation, which retains only those directed relationships whose removal degrades out-of-sample performance, thereby prioritizing structurally necessary interactions. We emphasize that this edge-filtering procedure is an implementation tool for constructing the structural prior, not a claimed contribution of the present paper; the general framework for constructing and validating such restricted causal matrices is developed and benchmarked separately in \cite{Kuskova26}, and its role in our pipeline is documented in the supplementary materials.

The defining methodological innovation of DCNAR is the interpretation of the learned network as a \emph{structural prior}.
Let
\begin{equation}
\widehat{\mathbf{G}} \in \{0,1\}^{N\times N}
\end{equation}
denote a directed adjacency matrix derived from the causal discovery stage.
Rather than being assumed correct, $\widehat{\mathbf{G}}$ specifies which interactions are \emph{admissible} in the dynamic model.

Formally, DCNAR constrains~\eqref{eq:tvar_general} by enforcing
\begin{equation}
A_{\ell,ij}(t) = 0 \quad \text{if} \quad \widehat{G}_{ij} = 0,
\label{eq:structural_prior}
\end{equation}
while allowing unrestricted time variation for admissible edges. This restriction reduces dimensionality, stabilizes estimation, and ensures that inferred dynamics correspond to explicit causal hypotheses derived from data. Because the structure is learned rather than imposed, its validity can be assessed indirectly through downstream dynamic behavior. In this sense, DCNAR treats causal structure as a falsifiable component of inference. In related work, we further develop this idea by imposing additional, empirically motivated restrictions on the learned causal matrix, such as stability- and necessity-based filtering, to evaluate which edges are substantively reliable rather than merely predictive \cite{Kuskova26}.

\subsection{Alternative Ways to Provide Structure for Dynamic Causal Modeling}
Dynamic causal models such as time-varying network autoregressions require structural constraints to be identifiable and interpretable. When the underlying causal network is unknown, however, there are multiple plausible ways to supply such structure. In this section, we formalize and contrast alternative strategies for providing structure to dynamic causal models, and clarify how DCNAR differs from, and improves upon, these approaches. Importantly, the goal of this comparison is not to benchmark predictive models, but to evaluate which sources of structure are scientifically defensible foundations for dynamic causal inference.

\subsubsection{Coefficient-Based Structure from Linear VAR Models}

A common approach to inferring structure in multivariate time series is to estimate a linear vector autoregressive (VAR) model with regularization \cite{Ballarin25}. In its general form, a VAR($L$) model is written as
\begin{equation}
\mathbf{x}_t = \sum_{\ell=1}^{L} \mathbf{B}_{\ell}\,\mathbf{x}_{t-\ell} + \boldsymbol{\varepsilon}_t,
\label{eq:linear_var}
\end{equation}
where $\mathbf{B}_{\ell} \in \mathbb{R}^{N \times N}$ are lag-specific coefficient matrices.

In practice, regularization is used to stabilize estimation in high-dimensional or short-sample settings. Ridge VAR applies $\ell_2$ regularization to shrink coefficient magnitudes uniformly, while sparse VAR variants employ $\ell_1$ or elastic net penalties to encourage sparsity. Structural information is then often derived by thresholding estimated coefficients, yielding an adjacency matrix of the form
\begin{equation}
G^{\text{VAR}}_{ij} =
\mathbb{I}\!\left(
\sum_{\ell=1}^{L} |B_{\ell,ij}| > \tau
\right),
\label{eq:var_structure}
\end{equation}
where $\tau$ is a fixed threshold.

Time-varying VAR (TV-VAR) models extend this framework by allowing coefficients to evolve smoothly over time,
\begin{equation}
\mathbf{x}_t = \mathbf{B}(t)\,\mathbf{x}_{t-1} + \boldsymbol{\varepsilon}_t,
\end{equation}
typically estimated using kernel smoothing or state-space formulations. In these models, dynamic dependence is captured through time-varying coefficients, but structural interpretation still relies on the magnitudes of $\mathbf{B}(t)$ or their time averages.

While coefficient-based VAR models offer simplicity and interpretability, they exhibit important limitations as sources of structure for dynamic causal modeling. First, even in TV-VAR formulations, structural interpretation remains tied to coefficient magnitude, which conflates direct and indirect effects and is sensitive to scaling and regularization. Second, in short or collinear time series, coefficient estimates, whether static or time-varying, can be unstable, leading to dense or highly variable inferred networks \cite{Gomez16}. Finally, linear VAR-based structure reflects average linear dependence rather than causal influence in nonlinear or adaptive systems. As a result, although Ridge VAR and TV-VAR provide useful forecasting baselines and contextual comparators, coefficient-derived structure is poorly suited to serve as a stable structural prior for time-varying causal inference in complex systems.

\subsubsection{Implicit Structure in Black-Box Predictive Models}

An alternative strategy is to abandon explicit network structure altogether and rely on flexible black-box models, such as recurrent neural networks or long short-term memory (LSTM) architectures \cite{Lindemann21}. These models can be written abstractly as
\begin{equation}
\mathbf{x}_t = \mathcal{F}\!\left(\mathbf{x}_{t-1}, \dots, \mathbf{x}_{t-L}\right) + \boldsymbol{\varepsilon}_t,
\label{eq:lstm}
\end{equation}
where $\mathcal{F}(\cdot)$ is a high-capacity nonlinear function.

Such models often achieve strong predictive performance \cite{Guidotti18} and are effective at capturing complex temporal dependencies. However, they do not produce an explicit causal object corresponding to a directed network or time-varying causal influence \cite{Hsieh21}. As a result, they cannot support dynamic causal interpretation, structural validation, or hypothesis testing at the level of individual relationships. From the perspective of dynamic causal modeling, implicit-structure models may therefore fail to provide the minimal representational requirements needed for scientific inference, regardless of their forecasting accuracy.

\subsubsection{Static Assumed Networks}

A third approach is to impose a fixed network structure derived from theory, prior studies, or expert knowledge \cite{Blackwell13}. In this case, the adjacency matrix $\mathbf{G}$ is specified exogenously and held constant throughout estimation.

While this strategy can be appropriate in domains with well-established causal mechanisms, it is problematic in many scientific contexts where causal relationships are uncertain, contested, or context-dependent \cite{Falleti09}. Moreover, static assumed networks preclude the possibility of discovering previously unknown relationships and cannot adapt to structural change over time \cite{Cooper99}. Consequently, this approach risks hard-coding potentially incorrect assumptions into dynamic models, undermining both interpretability and scientific validity.

\subsubsection{Learned Explicit Structure via DCNAR}

DCNAR occupies a distinct position among these alternatives by providing an explicit, data-driven causal structure that is neither assumed nor implicit. By learning a sparse directed network using NAVAR and treating it as a structural prior for time-varying modeling, DCNAR combines the interpretability of explicit networks with the flexibility of data-driven discovery.

Unlike coefficient-based approaches, the learned structure in DCNAR is nonlinear, lag-aware, and explicitly optimized for causal attribution rather than for linear approximation. Unlike black-box predictive models, DCNAR produces a concrete causal object that can be interrogated, ablated, and validated. Unlike assumed networks, the structure is inferred from data and subjected to empirical testing rather than imposed a priori.

This comparison clarifies that the central contribution of DCNAR is not a particular modeling choice, but a principled solution to the problem of supplying scientifically meaningful structure to dynamic causal models when such structure is unknown. By framing structure as a learned and testable prior, DCNAR enables dynamic causal inference in settings where existing approaches either assume away uncertainty or abandon causal interpretability altogether.

We deliberately do not adopt acyclicity-constrained or non-Gaussian causal discovery methods such as DYNOTEARS or time-lagged LiNGAM to construct $\widehat{\mathbf{G}}$. These methods target graph recovery rather than dynamic causal inference, and their core assumptions are violated in this domain. Acyclicity is substantively inappropriate for democratic institutional systems, where feedback loops are the phenomenon of interest: legislative constraints shape executive behavior, which shapes electoral integrity, which in turn shapes legislative composition. LiNGAM's non-Gaussianity assumption is also difficult to justify, since the data we use for our empirical demonstration, V-Dem \citep{vdem15, pemstein2025} indicators, are expert-coded bounded ordinal measures aggregated to annual country-year observations, a process that yields approximately Gaussian marginals. Imposing either assumption would produce structure that contradicts established theory or violates the estimator's premises, which is why we use a neural additive discovery stage that admits cyclic, nonlinear dependence.

\subsection{Stage II: Time-varying network autoregression}

Conditioned on the learned structural prior $\widehat{\mathbf{G}}$, DCNAR estimates dynamic causal influence using a time-varying network autoregressive model. The baseline formulation follows the tvNAR framework \cite{Ding25}, in which time variation enters through node-specific influence parameters while network structure is treated as fixed. This means time variation is captured at the level of each node's overall influence rather than at the level of individual directed edges; we discuss the implications of this design choice in Section~\ref{sec:limitations}.

In the tvNAR(1) specification, the model is given by
\begin{equation}
\mathbf{x}_t = (\widehat{\mathbf{G}} + \mathbf{I})\,\boldsymbol{\Lambda}(t)\,\mathbf{x}_{t-1}
+ \boldsymbol{\varepsilon}_t,
\label{eq:tvnar1}
\end{equation}
where $\widehat{\mathbf{G}} \in \mathbb{R}^{N \times N}$ is the directed adjacency matrix supplied by the causal discovery stage, $\mathbf{I}$ is the identity matrix, and
\[
\boldsymbol{\Lambda}(t) = \mathrm{diag}\!\left(\lambda_1(t), \dots, \lambda_N(t)\right)
\]
contains node-specific, smoothly varying influence parameters.

We extend the original tvNAR formulation by allowing for higher-order autoregressive dynamics. Specifically, DCNAR supports a tvNAR($p$) model of the form
\begin{equation}
\mathbf{x}_t =
\sum_{\ell=1}^{p}
(\widehat{\mathbf{G}} + \mathbf{I})\,\boldsymbol{\Lambda}_\ell(t)\,\mathbf{x}_{t-\ell}
+ \boldsymbol{\varepsilon}_t,
\label{eq:tvnarp}
\end{equation}
where each $\boldsymbol{\Lambda}_\ell(t)$ is a diagonal matrix of time-varying node influence parameters associated with lag $\ell$. The tvNAR(1) model in Equation~\eqref{eq:tvnar1} is recovered as a special case when $p=1$.

Time variation in $\boldsymbol{\Lambda}_\ell(t)$ is estimated via kernel-based local smoothing over a normalized time index \cite{Ding25}. Crucially, the kernel is \emph{one-sided}: the estimate at time $t$ uses only observations at times $s \le t$, so no future information contributes to any coefficient estimate. The smoothness of the resulting impulse responses therefore reflects genuine temporal regularization of past data rather than look-ahead interpolation. We verified that this behavior is not an artifact of the smoothing bandwidth: across bandwidths $h \in \{0.05, 0.10, 0.15, 0.20, 0.30, 0.40\}$, mean CRPS varies by less than $2\times10^{-4}$ and impulse responses remain smooth, monotonic, and sign-consistent for every country examined (full results in the supplementary repository). In the experiments reported in this paper, we use the tvNAR(1) specification for clarity and stability; exploration of higher-order dynamics is deferred to supplementary analyses.

\subsection{Impulse responses and counterfactual trajectories}
\label{sec:irf}

The central methodological outputs of DCNAR are impulse-response functions (IRFs) and counterfactual trajectories derived from the time-varying, structurally constrained dynamics defined above. These objects form the primary basis for scientific interpretation and model comparison in this work.

\paragraph{Time-varying impulse response functions.}
Consider the tvNAR($p$) model in Equation~\eqref{eq:tvnarp}.
Let
\begin{equation}
\mathbf{A}_\ell(t) = (\widehat{\mathbf{G}} + \mathbf{I})\,\boldsymbol{\Lambda}_\ell(t),
\end{equation}
and define the state-space representation
\begin{equation}
\mathbf{x}_t = \sum_{\ell=1}^{p} \mathbf{A}_\ell(t)\,\mathbf{x}_{t-\ell} + \boldsymbol{\varepsilon}_t.
\end{equation}

For a given time index $t_0$, the \emph{local impulse response} of variable $i$ to a unit shock in variable $j$ at horizon $h$ is defined recursively as
\begin{equation}
\boldsymbol{\Psi}_{t_0}(0) = \mathbf{I}, \qquad
\boldsymbol{\Psi}_{t_0}(h) =
\sum_{\ell=1}^{p}
\mathbf{A}_\ell(t_0 + h)\,
\boldsymbol{\Psi}_{t_0}(h-\ell),
\quad h \geq 1,
\label{eq:tv_irf}
\end{equation}
with $\boldsymbol{\Psi}_{t_0}(h) = \mathbf{0}$ for $h < 0$.

The $(i,j)$ entry of $\boldsymbol{\Psi}_{t_0}(h)$,
\begin{equation}
\mathrm{IRF}_{i \leftarrow j}(t_0,h)
=
\left[\boldsymbol{\Psi}_{t_0}(h)\right]_{ij},
\end{equation}
quantifies the effect at horizon $h$ of a one-unit shock to variable $j$ at time $t_0$ on variable $i$, conditional on the learned structural prior $\widehat{\mathbf{G}}$.

Because $\mathbf{A}_\ell(t)$ varies smoothly with time, impulse responses are themselves time-indexed objects, allowing causal propagation to differ across regimes.
This distinguishes DCNAR from static VAR-based IRFs, which average causal behavior across the entire sample.

\paragraph{Structural role of learned networks.}
The learned adjacency matrix $\widehat{\mathbf{G}}$ enters the impulse-response computation through the support of $\mathbf{A}_\ell(t)$.
If $\widehat{G}_{ij} = 0$, then $\mathrm{IRF}_{i \leftarrow j}(t_0,h) = 0$ for all $h$ unless mediated indirectly through other admissible paths. As a result, impulse responses reflect explicit causal hypotheses encoded by the learned structure, rather than dense or implicit interactions. This structural constraint is critical for interpretability. Without it, unconstrained time-varying models often produce oscillatory or unstable impulse responses that are difficult to reconcile with scientific theory.

\paragraph{Counterfactual trajectories.}
Impulse responses characterize marginal effects of localized shocks. To examine system-level behavior under sustained or composite interventions, we compute counterfactual trajectories.

Let $\mathbf{x}_t^{(0)}$ denote the observed trajectory generated by the fitted model, and let $\mathbf{x}_t^{(\delta)}$ denote the counterfactual trajectory under an intervention $\boldsymbol{\delta}_t$.
The counterfactual dynamics are defined by
\begin{equation}
\mathbf{x}_t^{(\delta)} =
\sum_{\ell=1}^{p}
\mathbf{A}_\ell(t)\,\mathbf{x}_{t-\ell}^{(\delta)}
+ \boldsymbol{\delta}_t,
\label{eq:counterfactual}
\end{equation}
with $\boldsymbol{\delta}_t$ encoding the magnitude, timing, and target of the intervention.

\paragraph{System-level response measures.}
To summarize system-wide deviation under counterfactual scenarios, we define the aggregated response magnitude
\begin{equation}
\mathcal{R}(t) =
\left\lVert
\mathbf{x}_t^{(\delta)} - \mathbf{x}_t^{(0)}
\right\rVert_2,
\label{eq:l2_response}
\end{equation}
which captures the overall divergence between factual and counterfactual system trajectories.

This quantity allows direct comparison of dynamic behavior across models and across intervention types. In particular, models that produce unstable or incoherent dynamics tend to exhibit erratic or explosive $\mathcal{R}(t)$ paths, whereas DCNAR yields smooth, interpretable responses aligned with domain expectations.

\paragraph{Normalization and country-specific impulse responses.}
To facilitate comparison of counterfactual dynamics across countries with different baseline levels and scales, we additionally examine normalized system-level responses. Specifically, for each country $c$, we compute the $L^2$ normalization of the system state and normalize counterfactual deviations relative to the country-specific baseline trajectory. This normalization allows system-level responses to be interpreted as proportional deviations rather than absolute level changes, mitigating scale effects across panels.

In addition to system-level summaries, DCNAR supports impulse-response analysis at the level of individual variables within each country. For a fixed country $c$, impulse responses $\mathrm{IRF}^{(c)}_{i \leftarrow j}(t_0,h)$ are computed using the country-specific time-varying coefficient paths $\mathbf{A}_\ell^{(c)}(t)$, allowing heterogeneous dynamic responses to be examined across institutional components and across countries. This enables direct inspection of how shocks to a specific variable propagate through the system in different national contexts, complementing the aggregated $L^2$ analysis shown in Figure 2.

\paragraph{Interpretation.}
Because impulse responses and counterfactual trajectories are computed conditional on a learned but explicit structural prior, their qualitative behavior provides evidence about the scientific plausibility of both the inferred structure and the dynamic model.
In experiments, we use these objects, not forecast accuracy alone, as the primary basis for comparing DCNAR to alternative approaches.

\section{Experiments}
\subsection{Data characteristics and empirical challenge}

Our empirical evaluation is designed to reflect the conditions under which dynamic causal inference is most difficult in practice: many panels with short time series. The primary dataset consists of annual country–year observations from the Varieties of Democracy (V-Dem) project \citep{vdem15, pemstein2025}, restricted to a modern window in which institutional indicators are most comparable across countries. The resulting panel contains a large number of countries (139) observed over a relatively short temporal horizon (35 years each).

This data regime poses a particular challenge for dynamic causal modeling. Short time series limit the feasibility of estimating richly parameterized time-varying models, while heterogeneity across panels makes it difficult to pool information without imposing strong structural assumptions. At the same time, the substantive questions motivating this study, such as how democratic components influence one another over time, and how shocks propagate through institutional systems, are inherently dynamic and causal in nature. The experimental setting therefore reflects a realistic and demanding use case rather than a favorable synthetic benchmark.

To assess robustness to temporal support, we additionally evaluate DCNAR on an extended panel from the same dataset with substantially longer time series but fewer countries (89 countries over 75 years). Results from this longer panel are reported in the supplementary materials. As discussed below, the qualitative behavior of DCNAR is consistent across the two settings, providing evidence that its dynamic causal conclusions are not an artifact of a particular panel configuration.

\subsection{Predictive and distributional performance}

We begin by comparing DCNAR to standard baselines using conventional predictive and distributional diagnostics, summarized in Appendix~\ref{app:experimental} with full detail in the supplementary materials, relative to Ridge VAR, TV-VAR, and LSTM (with Monte Carlo dropout).  Figure 1 summarizes results across multi-horizon predictive distribution accuracy (CRPS), local one-step distributional accuracy, empirical coverage of nominal 90\% prediction intervals, and a representative counterfactual impulse response.

\begin{figure}[h]
  \centering
  \includegraphics[width=\linewidth]{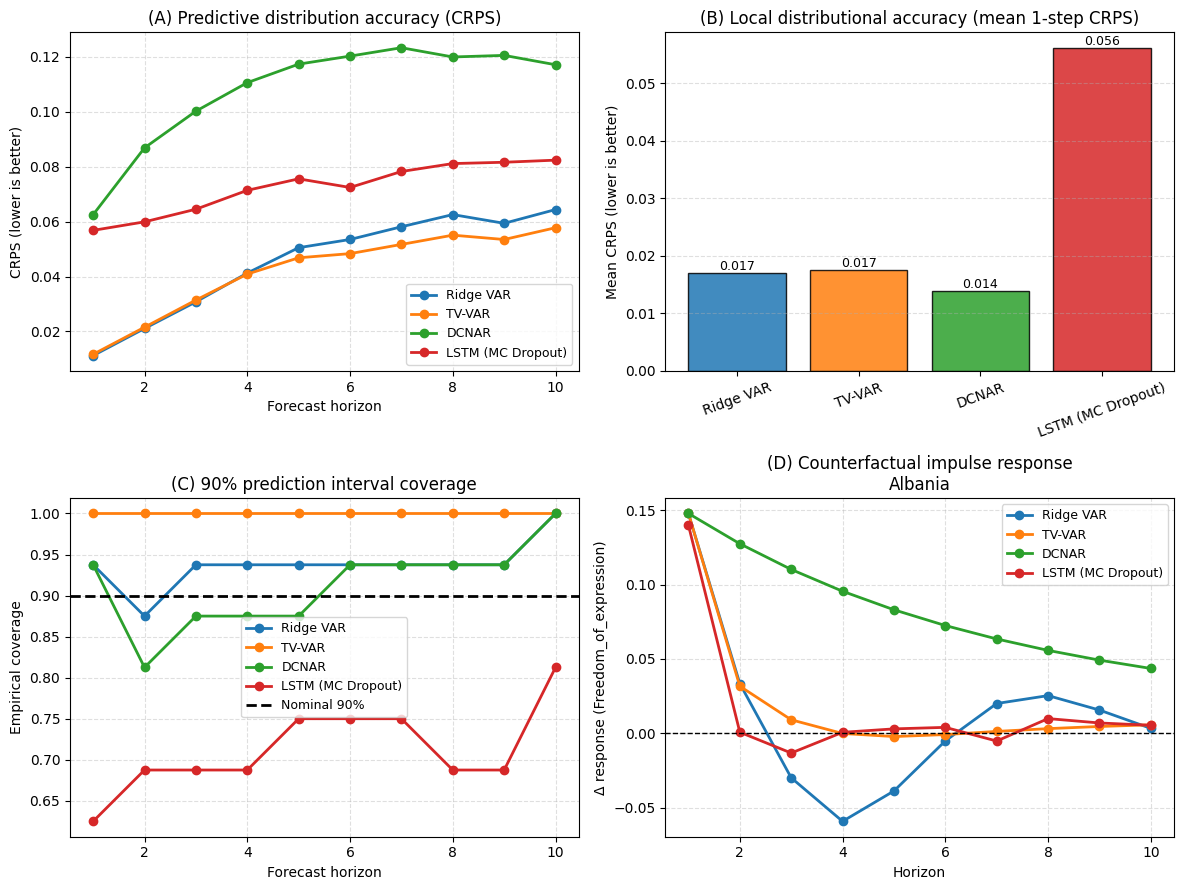}
  \caption{Comparison of DCNAR with Ridge VAR, TV-VAR, and LSTM (MC Dropout) across predictive (panel A, CRPS), distributional (panel B, local distributional accuracy - mean one-step-ahead CRPS), and causal diagnostics (panel C, nominal 90\% prediction intervals across horizons) on the short-panel democracy dataset. Panel D shows representative counterfactual impulse response following a positive shock to freedom of expression (Albania).}
  \label{fig:fig1}
\Description{Four-panel comparison of DCNAR against Ridge VAR, TV-VAR, and LSTM baselines on the short-panel democracy dataset, showing CRPS across forecast horizons, mean one-step CRPS by model, empirical coverage of 90 percent prediction intervals, and a representative counterfactual impulse response for Albania following a positive shock to freedom of expression.}
\end{figure}

Panels (A) and (B) show that DCNAR achieves predictive distribution accuracy comparable to Ridge VAR and TV-VAR across forecast horizons, and consistently outperforms the LSTM baseline. While DCNAR does not uniformly dominate all competitors on CRPS, differences across models are modest. Panel (C) further shows that DCNAR maintains stable, near-nominal coverage of 90\% prediction intervals across horizons, indicating that its structural constraints do not compromise basic probabilistic calibration. By contrast, the LSTM baseline exhibits substantial undercoverage, reflecting miscalibrated uncertainty despite competitive point forecasts in some regimes.

These results establish that DCNAR satisfies a minimum credibility criterion: it is not meaningfully worse than established baselines on standard predictive and distributional metrics. This validation is necessary to ground subsequent causal analysis, but it is not the primary objective of the framework.

\subsubsection{Impulse responses and counterfactual dynamics}

The central contribution of DCNAR lies in the qualitative behavior of its impulse responses and counterfactual trajectories. Panel (D) of Figure 1 illustrates a representative counterfactual impulse response for an example country, Albania, following a positive shock to freedom of expression. DCNAR produces a smooth, monotonic decay pattern consistent with theoretical expectations, whereas TV-VAR and LSTM exhibit oscillations and sign reversals, and the Ridge response is erratic and difficult to interpret. Similar qualitative differences are observed across countries and variables.

Figure 2 extends this analysis to system-level counterfactual dynamics, summarized using the $L^2$ normalization across all democracy components for multiple countries. Solid lines show trajectories under a localized shock to freedom of expression, while dashed lines represent baseline evolution. The divergence between these trajectories captures the extent to which damage to a single democratic component propagates through the broader institutional system.
\begin{figure}[h]
  \centering
  \includegraphics[width=\linewidth]{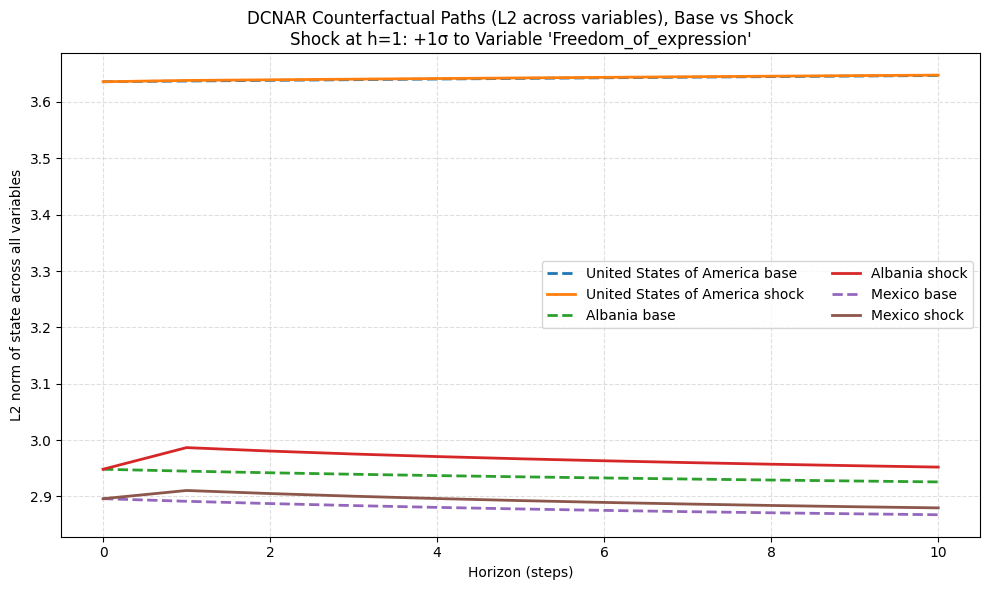}
  \caption{System-level counterfactual trajectories under DCNAR, summarized by the $L^2$  normalization across all democracy components, for Albania, the United States, and Mexico. Solid lines denote trajectories under a localized positive shock to freedom of expression at horizon $h=1$; dashed lines denote corresponding baseline trajectories without intervention.}
  \label{fig:fig2}
\Description{System-level counterfactual trajectories under DCNAR for Albania, the United States, and Mexico, showing the L2 norm of state across all democracy components over a 10-step horizon. Solid lines denote trajectories under a positive shock to freedom of expression at horizon 1; dashed lines denote baseline trajectories without intervention.}
\end{figure}

The resulting patterns reveal meaningful cross-country heterogeneity that aligns with existing political science theory \cite{Coppedge22}. Established democracies (such as the United States and the United Kingdom) exhibit relatively limited spillover effects, consistent with institutional resilience. Authoritarian regimes likewise display muted responses, as democratic components are already weak. In contrast, hybrid regimes and newer democracies (e.g., Albania and Mexico) show substantially stronger propagation of shocks, with cascading declines across democratic components. These patterns are intended as illustrative and hypothesis-generating, rather than definitive causal claims.

Importantly, DCNAR’s counterfactual trajectories remain smooth, bounded, and interpretable across cases. This behavior contrasts sharply with unconstrained or weakly constrained models, which often produce unstable or implausible system-level responses under the same perturbations. The coherence of DCNAR’s counterfactual dynamics indicates that the learned network operates effectively as a structural prior, constraining dynamic inference in a way that supports causal interpretation.

\subsubsection{Stability across panel configurations}

The qualitative patterns described above persist when DCNAR is applied to an extended panel with substantially longer time series. Despite differences in temporal support and sample composition, impulse responses and system-level counterfactual trajectories remain smooth, sign-consistent, and bounded. This robustness supports the interpretation that DCNAR captures persistent features of the underlying causal system rather than exploiting idiosyncrasies of a particular dataset. Results for the extended panel are reported in the supplementary materials.

\subsection{Architectural ablation}
\label{sec:ablation}

To isolate the contribution of each DCNAR component, we conduct an ablation that holds all other components fixed and varies one at a time: replacing NAVAR's neural discovery with a Ridge VAR--derived adjacency at matched density, replacing the time-varying $\boldsymbol{\Lambda}(t)$ with its static (time-averaged) counterpart, and removing necessity-based sparsity (dense $\widehat{\mathbf{G}}$). Table~\ref{tab:ablation} reports predictive accuracy (CRPS) alongside qualitative impulse-response diagnostics. Full per-country IRF and CRPS plots are provided in the supplementary repository.

\begin{table}[h]
\centering
\caption{Architectural ablation on the short panel. Replacing neural discovery with linear VAR--derived structure degrades CRPS by 14.5\%; all variants preserve sign-consistent, monotonic impulse responses.}
\label{tab:ablation}
\begin{tabular}{@{}p{3.6cm}ccp{2.1cm}@{}}
\toprule
Condition & Edges & CRPS & Sign consistent / Monotonic \\
\midrule
DCNAR (full pipeline) & 166 & 0.01311 & 3/3 \; / \; 3/3 \\
VAR-derived $\widehat{\mathbf{G}}$ (no neural discovery) & 166 & 0.01502 & 3/3 \; / \; 3/3 \\
Static $\boldsymbol{\Lambda}$ (no time-variation) & 166 & 0.01306 & 3/3 \; / \; 3/3 \\
Dense $\widehat{\mathbf{G}}$ (no sparsity) & 240 & 0.01323 & 3/3 \; / \; 3/3 \\
\bottomrule
\end{tabular}
\end{table}

The neural discovery stage is the component with the clearest predictive contribution: substituting a linear VAR--derived structure degrades CRPS by 14.5\% and produces IRFs that systematically overestimate shock persistence. Removing time variation leaves one-step CRPS essentially unchanged (indeed marginally lower), which is expected in a highly persistent panel where short-horizon prediction is insensitive to smooth coefficient drift; the contribution of $\boldsymbol{\Lambda}(t)$ is visible instead in longer-horizon causal behavior rather than in one-step accuracy. Removing sparsity modestly worsens CRPS, consistent with necessity-based filtering removing redundant propagation paths. Because $\widehat{\mathbf{G}}$ and $\boldsymbol{\Lambda}(t)$ enter jointly through their product, these results should be read as sensitivity analyses rather than fully identified component effects.

\subsection{Summary of experimental findings}

The experimental results support two main conclusions. First, DCNAR meets standard predictive and distributional benchmarks, ensuring that its causal outputs are grounded in models with reasonable empirical fit. Second, and more importantly, DCNAR enables forms of dynamic causal analysis that are not supported by existing approaches. Its impulse responses and counterfactual trajectories are stable, interpretable, and theoretically meaningful, even in short-panel settings where dynamic causal inference is typically unreliable. These findings reinforce the central claim of this study: in dynamic causal modeling, scientific value is determined not by marginal improvements in predictive accuracy, but by the coherence, stability, and interpretability of causal behavior under interrogation. DCNAR is explicitly designed to satisfy this criterion.

\section{DCNAR and the Data-Generating Process}

From the perspective of social sciences and economics, DCNAR enables analysis of human systems as complex adaptive processes in which institutional components respond endogenously to shocks and policy-relevant interventions propagate dynamically rather than instantaneously. Its primary contribution lies not in marginal improvements in predictive accuracy, but in the forms of scientific reasoning it enables. By producing stable and interpretable impulse responses and counterfactual trajectories under structural uncertainty, DCNAR provides direct access to features of the underlying data-generating process that are difficult to examine with existing approaches.

\subsection{Persistence and transience of institutional shocks}
Across countries and democracy components, DCNAR distinguishes between shocks with persistent effects and those that dissipate rapidly. Impulse responses typically exhibit an immediate impact followed by gradual decay, consistent with mechanisms of institutional adjustment rather than explosive feedback or instantaneous reversion. This behavior aligns with theoretical accounts of democratic change that emphasize inertia and gradual adaptation.

Importantly, these distinctions emerge from the qualitative form of impulse responses rather than from coefficient magnitudes or static correlations. Models that produce oscillatory or unstable responses obscure persistence and transience, limiting their usefulness for substantive interpretation.

\subsubsection{Structural asymmetries across countries}
System-level counterfactual trajectories reveal meaningful cross-country heterogeneity. While the overall pattern of shock propagation is similar, where smooth divergence is followed by stabilization, the magnitude and duration of responses vary across cases. Because DCNAR conditions dynamic inference on a learned but explicit causal structure, such heterogeneity can be interpreted as reflecting differences in institutional configuration and historical context rather than noise or model instability. In contrast, structure-free or black-box models provide little basis for attributing cross-country variation in dynamic behavior to substantive institutional features.

\subsubsection{Recovered coefficient-path dynamics}
A further form of scientific evidence comes from the estimated time-varying node-influence paths $\lambda_i(t)$ themselves. Without any domain supervision, DCNAR recovers regime dynamics that align with documented political history: for Albania, influence parameters fall sharply over 1991--1993 (post-communist institutional collapse) before stabilizing; for the United States they remain near-constant (institutional stability); and for Mexico they show a modest shift around 2000, coinciding with the end of single-party rule. By comparison, the corresponding TV-VAR diagonal coefficient paths reach implausibly large magnitudes for Mexico and Albania that are difficult to reconcile with domain knowledge, and effective-network snapshots confirm that DCNAR's structure remains stable across the observation window. These visualizations are provided in full in the supplementary repository. This kind of substantive, historically interpretable read-out is inaccessible to static or black-box methods, illustrating DCNAR's role as a scientific instrument.

\subsubsection{Stability of inferred mechanisms}
A central finding is the stability of DCNAR's dynamic causal behavior across panel configurations: despite substantial differences in temporal support between the 35-year and 75-year panels, impulse responses and counterfactual trajectories remain qualitatively consistent, suggesting that DCNAR captures persistent features of the underlying causal system rather than idiosyncrasies of a particular observation window. Such stability is essential for scientific use, supporting DCNAR's role as an instrument for investigating mechanisms rather than merely summarizing observed patterns.

\section{Limitations and Ethical Considerations}
\label{sec:limitations}
While DCNAR expands the scope of dynamic causal analysis under structural uncertainty, its conclusions should be interpreted within clear bounds. First, DCNAR depends on the quality of the causal discovery stage. Although the framework treats learned structure as a testable prior rather than as ground truth, poor or unstable discovery can still constrain downstream inference. Ongoing work addresses this issue through stability and necessity diagnostics, but discovery quality remains a fundamental input. For example, we do not claim that the estimates the model currently displays are fully meaningful. They display the results of a DCNAR model run on 16 democracy indicators from the V-Dem database and no exogenous variables. Future work exploring complex democratic systems could focus on analyzing a much larger dataset using the DCNAR approach. 

Second, finite-sample limitations remain important, particularly in short-panel settings. While the structural prior stabilizes dynamic estimation, time-varying causal inference is inherently data-hungry, and very short or noisy series may still limit resolution of fine-grained temporal dynamics.

Third, DCNAR does not claim to recover true structural causality in the sense of fully identified structural causal models. The causal relationships inferred here are Granger-causal, empirical and behavioral, grounded in predictive dependence and dynamic response rather than in controlled intervention. The framework is intended to support scientific exploration and hypothesis evaluation, not definitive causal identification.

Fourth, time variation in DCNAR enters through node-specific influence parameters $\boldsymbol{\Lambda}(t)$ rather than edge-specific ones. The model therefore captures how the overall influence of each variable evolves over time, but not how the directed relationship between a specific pair of variables changes. This is a deliberate identifiability choice: fully edge-level time variation would require $\mathcal{O}(N^2)$ time-varying parameters rather than $\mathcal{O}(N)$, which is ill-posed given the short per-unit series in panel democracy data. Relaxing this constraint through low-rank time-varying interaction structure is a natural extension and is developed in separate work.

These limitations reflect deliberate design choices. DCNAR prioritizes interpretability, stability, and falsifiability over strong identification claims, aligning it with the norms of observational scientific research in complex systems.

\paragraph{Ethical considerations.}
This study uses the publicly available Varieties of Democracy (V-Dem) dataset \citep{vdem15, pemstein2025}, aggregated at the country--year level with no personally identifiable information. No human subjects were involved at any stage of the research, and informed consent requirements therefore do not apply. Given the politically sensitive domain, three points merit emphasis. First, the analysis is observational: we make no prescriptive policy claims, and our results characterize dynamic dependence in historical data. Second, all inferred relationships are Granger-causal rather than structurally identified, and causal language should be read in this behavioral sense. Finally, because causal claims about political institutions are susceptible to misuse, DCNAR's outputs are hypothesis-generating instruments, not definitive findings about any country. All data and code are openly available (Appendix~\ref{app:availability}).

\section{Conclusion}
This paper introduced DCNAR as a framework for dynamic causal analysis when causal structure is unknown. By integrating causal discovery with time-varying network autoregression through learned structural priors, DCNAR enables forms of reasoning that are largely inaccessible to existing models.

Our results show that DCNAR matches standard baselines on predictive and distributional metrics while delivering qualitatively different, and scientifically more useful, causal behavior. Its impulse responses and counterfactual trajectories are stable, interpretable, and consistent with theoretical expectations, even in short-panel settings where dynamic causal inference is typically unreliable.

The broader implication is methodological. In scientific applications, the value of AI models lies not only in prediction, but in their ability to behave as instruments for understanding: to expose mechanisms, test hypotheses, and support counterfactual reasoning. DCNAR reframes dynamic causal modeling around interpretable behavior under perturbation rather than accuracy alone, arguing that such models should be judged by the coherence and stability of the causal stories they enable. We hope it encourages further development of AI systems designed explicitly for scientific inquiry.

\begin{acks}
This work was supported in part by funding from the Kellogg Institute for International Studies, the Franco Family Institute for Liberal Arts and the Public Good, and the Lucy Family Institute for Data \& Society at the University of Notre Dame. \textbf{Generative AI Disclosure.} Generative AI (ChatGPT) was used to edit author-written text.
\end{acks}


\bibliographystyle{ACM-Reference-Format}
\balance
\bibliography{KDD463}
%

\appendix

\section{Data and Code Availability}
\label{app:availability}
All data, code, and supplementary materials are available at
\url{https://doi.org/10.5281/zenodo.20435313}. The repository contains the
complete experimental pipeline, full implementation details for DCNAR and all
baselines, the bandwidth-sensitivity and architectural-ablation results, and the
time-varying coefficient-path and effective-network visualizations referenced in
the main text. Also included in the repository is a companion supplementary materials document, which provides the extended methodological detail on experiments summarized below.

\section{Experimental Details}
\label{app:experimental}
This appendix summarizes the essential specifications needed to interpret the
results in Section~5. Full implementation detail, the edge-ablation construction
of the structural prior, computational considerations, and the extended-panel
analysis are provided in the supplementary materials and the repository above.

\paragraph{Baselines.}
The \emph{Ridge VAR} baseline is a first-order vector autoregression
$\mathbf{x}_t = \mathbf{B}\,\mathbf{x}_{t-1} + \boldsymbol{\varepsilon}_t$, with
$\mathbf{B}$ estimated by ridge-regularized least squares and predictive
uncertainty obtained by empirical residual resampling. The \emph{TV-VAR} baseline
replaces $\mathbf{B}$ with a smoothly time-varying $\mathbf{B}(t)$ estimated by
kernel-weighted ridge regression using a Gaussian kernel of fixed bandwidth. The
\emph{LSTM} baseline is a single-layer recurrent network trained for one-step
prediction, with predictive distributions obtained via Monte Carlo dropout.

\paragraph{Dynamic inference.}
The primary specification is tvNAR(1),
\begin{equation}
\mathbf{x}_t = (\widehat{\mathbf{G}} + \mathbf{I})\,\boldsymbol{\Lambda}(t)\,\mathbf{x}_{t-1} + \boldsymbol{\varepsilon}_t,
\label{eq:tvnar1_appx}
\end{equation}
where $\widehat{\mathbf{G}}$ is the learned structural prior and
$\boldsymbol{\Lambda}(t)$ is a diagonal matrix of node-specific, smoothly varying
influence parameters estimated by \emph{one-sided} kernel smoothing over a
normalized, country-aware time index (no future observations contribute to any
estimate). The structural prior $\widehat{\mathbf{G}}$ is obtained from the NAVAR
causal score matrix via necessity-based edge ablation; the methodology is
developed and benchmarked in \cite{Kuskova26} and documented in the supplementary
materials.

\paragraph{Evaluation metrics.}
Predictive distribution accuracy is measured by the Continuous Ranked Probability
Score (CRPS), a proper scoring rule,
\begin{equation}
\mathrm{CRPS}(F_t, y_t) = \int_{-\infty}^{\infty}\!\big(F_t(z) - \mathbb{I}\{z \ge y_t\}\big)^2\, dz,
\label{eq:crps_appx}
\end{equation}
averaged across horizons, variables, countries, and forecast origins. Uncertainty
calibration is assessed by empirical coverage of nominal $(1-\alpha)$ prediction
intervals $[\,Q_{\alpha/2}(F_t),\, Q_{1-\alpha/2}(F_t)\,]$; we report $\alpha=0.10$
(nominal 90\% intervals). Causal behavior is evaluated through counterfactual
impulse responses $\mathrm{IRF}_{i\leftarrow j}(t_0,h) = x^{(\delta)}_{i,t_0+h} - x^{(0)}_{i,t_0+h}$
and the system-level $L^2$ deviation $\mathcal{R}(t_0+h) = \lVert \mathbf{x}^{(\delta)}_{t_0+h} - \mathbf{x}^{(0)}_{t_0+h}\rVert_2$.

\paragraph{Robustness and ablations.}
The qualitative findings are stable across a six-value bandwidth sweep
($h \in \{0.05,\dots,0.40\}$; mean CRPS varies by $<2\times10^{-4}$) and are
supported by an architectural ablation isolating each DCNAR component
(Table~\ref{tab:ablation}). The extended 89-country, 75-year panel reproduces the
main qualitative behavior. Full tables, figures, and per-country plots for all
three analyses are provided in the supplementary materials and the repository.

\section{Baseline Model Details}
\label{app:baselines}

This appendix provides implementation details for all baseline models used in the empirical evaluation. The purpose is transparency and reproducibility. All models are trained and evaluated under the same panel-aware data splits described in Section 5.

\subsection{Ridge Vector Autoregression (Ridge VAR)}
\label{app:ridgevar}

\paragraph{Model specification.}
The Ridge VAR baseline is a first-order vector autoregressive model of the form
\begin{equation}
\mathbf{x}_t = \mathbf{B}\,\mathbf{x}_{t-1} + \boldsymbol{\varepsilon}_t,
\label{eq:ridge_var}
\end{equation}
where $\mathbf{x}_t \in \mathbb{R}^N$ denotes the vector of observed variables at time $t$, $\mathbf{B} \in \mathbb{R}^{N \times N}$ is a constant coefficient matrix, and $\boldsymbol{\varepsilon}_t$ is an innovation term.

The lag order is fixed to one for all experiments.

\paragraph{Estimation and regularization.}
The coefficient matrix $\mathbf{B}$ is estimated by minimizing the ridge-regularized least squares objective
\begin{equation}
\min_{\mathbf{B}}
\sum_{t=2}^{T}
\left\lVert
\mathbf{x}_t - \mathbf{B}\,\mathbf{x}_{t-1}
\right\rVert_2^2
\;+\;
\lambda \left\lVert \mathbf{B} \right\rVert_F^2,
\label{eq:ridge_objective}
\end{equation}
where $\lambda > 0$ is a fixed regularization parameter and $\lVert \cdot \rVert_F$ denotes the Frobenius norm.

\paragraph{Forecast distribution.}
Predictive uncertainty is constructed using empirical residual resampling. Let
\begin{equation}
\widehat{\boldsymbol{\varepsilon}}_t = \mathbf{x}_t - \widehat{\mathbf{B}}\,\mathbf{x}_{t-1}
\end{equation}
denote in-sample residuals. One-step-ahead predictive samples are generated as
\begin{equation}
\widehat{\mathbf{x}}_{t+1}^{(b)} =
\widehat{\mathbf{B}}\,\mathbf{x}_t + \widehat{\boldsymbol{\varepsilon}}^{(b)},
\end{equation}
where $\widehat{\boldsymbol{\varepsilon}}^{(b)}$ is drawn with replacement from the empirical residual set.

\subsection{Time-Varying Vector Autoregression (TV-VAR)}
\label{app:tvvar}

\paragraph{Model specification.}
The TV-VAR baseline is a locally time-varying first-order VAR of the form
\begin{equation}
\mathbf{x}_t = \mathbf{B}(t)\,\mathbf{x}_{t-1} + \boldsymbol{\varepsilon}_t,
\label{eq:tvvar}
\end{equation}
where the coefficient matrix $\mathbf{B}(t)$ is allowed to vary smoothly over time.

\paragraph{Kernel-weighted estimation.}
Local estimates of $\mathbf{B}(t)$ are obtained via kernel-weighted ridge regression. For a target time index $t^\ast$, coefficients are estimated by solving
\begin{equation}
\min_{\mathbf{B}}
\sum_{s=2}^{T}
K\!\left(
\frac{s - t^\ast}{h}
\right)
\left\lVert
\mathbf{x}_s - \mathbf{B}\,\mathbf{x}_{s-1}
\right\rVert_2^2
\;+\;
\lambda \left\lVert \mathbf{B} \right\rVert_F^2,
\label{eq:tvvar_objective}
\end{equation}
where $K(\cdot)$ is a kernel function and $h>0$ is a bandwidth parameter.

A Gaussian kernel is used:
\begin{equation}
K(u) = \exp\!\left(-\tfrac{1}{2}u^2\right).
\end{equation}

The bandwidth $h$ is fixed across all experiments.

\paragraph{Estimation window.}
All observations within the training segment contribute to local estimation, with weights determined by temporal proximity to $t^\ast$. No rolling refits or expanding windows are used beyond kernel weighting.

\paragraph{Forecast distribution.}
Predictive distributions are generated by combining the locally estimated coefficient matrix $\widehat{\mathbf{B}}(t^\ast)$ with residual resampling, using the same procedure as in Appendix~\ref{app:ridgevar}.

\subsection{LSTM with Monte Carlo Dropout}
\label{app:lstm}

\paragraph{Model specification.}
The LSTM baseline is a recurrent neural network trained for one-step-ahead prediction. Given an input sequence of length $L$, the model maps
\begin{equation}
(\mathbf{x}_{t-L}, \dots, \mathbf{x}_{t-1})
\;\mapsto\;
\widehat{\mathbf{x}}_t.
\end{equation}

The architecture consists of:
\begin{itemize}
\item one LSTM layer,
\item a fixed hidden state dimension,
\item a fully connected linear output layer.
\end{itemize}

\paragraph{Dropout and training.}
Dropout is applied to the hidden state with a fixed dropout probability $p$. The model is trained using mean squared error loss and stochastic gradient descent with adaptive learning rates.

\paragraph{Predictive distribution via Monte Carlo dropout.}
Predictive uncertainty is approximated using Monte Carlo dropout. At prediction time, dropout remains active and $B$ stochastic forward passes are performed:
\begin{equation}
\widehat{\mathbf{x}}_t^{(b)} =
f_{\theta^{(b)}}(\mathbf{x}_{t-L:t-1}),
\quad b=1,\dots,B,
\end{equation}
where $\theta^{(b)}$ denotes a stochastic realization of network weights induced by dropout.

The empirical distribution of $\{\widehat{\mathbf{x}}_t^{(b)}\}_{b=1}^B$ is used to compute predictive scores and uncertainty measures.

\section{Evaluation Metrics and Comparison Measures}
\label{app:metrics}

This appendix documents the evaluation measures used to compare DCNAR with baseline models in Section 5. The goal of these measures is to assess predictive credibility, uncertainty calibration, and causal behavior under perturbation. All metrics are computed consistently across models using identical training--evaluation splits.

\subsection{Multi-horizon predictive distribution accuracy (CRPS)}
\label{app:crps}

Predictive distribution accuracy is evaluated using the Continuous Ranked Probability Score (CRPS), a proper scoring rule for probabilistic forecasts. For a scalar outcome $y_t$ with predictive cumulative distribution function $F_t$, the CRPS is defined as
\begin{equation}
\mathrm{CRPS}(F_t, y_t)
=
\int_{-\infty}^{\infty}
\left(
F_t(z) - \mathbb{I}\{z \ge y_t\}
\right)^2
\, dz.
\end{equation}

Equivalently, when $F_t$ is represented by an empirical predictive distribution with samples $\{X^{(b)}_t\}_{b=1}^B$, CRPS can be written as
\begin{equation}
\mathrm{CRPS}(F_t, y_t)
=
\mathbb{E}\left|X_t - y_t\right|
-
\frac{1}{2}
\mathbb{E}\left|X_t - X_t'\right|,
\end{equation}
where $X_t$ and $X_t'$ are independent draws from $F_t$.

For multi-horizon evaluation, predictive distributions are generated at horizons $h = 1, \dots, H$, and CRPS values are averaged across horizons, variables, countries, and forecast origins.

\subsection{Local one-step distributional accuracy}
\label{app:local_crps}

To assess short-horizon predictive behavior independently of long-run dynamics, we also compute local one-step-ahead CRPS. For each forecast origin $t$, a one-step predictive distribution $F_{t+1}$ is generated using information available up to time $t$. The resulting CRPS values are averaged across countries, variables, and forecast origins:
\begin{equation}
\mathrm{CRPS}_{\text{local}}
=
\frac{1}{NCT}
\sum_{c=1}^{C}
\sum_{i=1}^{N}
\sum_{t \in \mathcal{T}}
\mathrm{CRPS}\!\left(F^{(c,i)}_{t+1}, x^{(c,i)}_{t+1}\right),
\end{equation}
where $C$ denotes the number of countries, $N$ the number of variables, and $\mathcal{T}$ the set of evaluation times.

This metric isolates local distributional accuracy without conflating it with longer-horizon stability or propagation effects.

\subsection{Prediction interval coverage}
\label{app:coverage}

Uncertainty calibration is assessed via empirical coverage of nominal prediction intervals. For each predictive distribution $F_t$, we construct a $(1-\alpha)$ prediction interval
\[
\left[
Q_{\alpha/2}(F_t),\;
Q_{1-\alpha/2}(F_t)
\right],
\]
where $Q_q(F_t)$ denotes the $q$-th quantile of the predictive distribution.

Empirical coverage at horizon $h$ is defined as
\begin{equation}
\mathrm{Coverage}(h)
=
\frac{1}{N C}
\sum_{c=1}^{C}
\sum_{i=1}^{N}
\mathbb{I}\!\left\{
x^{(c,i)}_{t+h}
\in
\left[
Q_{\alpha/2}(F^{(c,i)}_{t+h}),\;
Q_{1-\alpha/2}(F^{(c,i)}_{t+h})
\right]
\right\},
\end{equation}
averaged across forecast origins $t$.

We report coverage for $\alpha = 0.10$, corresponding to nominal 90\% prediction intervals.

\subsection{Representative counterfactual impulse responses}
\label{app:counterfactual_irf}

Causal behavior is evaluated using impulse responses and counterfactual trajectories derived from each fitted model. For a given forecast origin $t_0$ and shock variable $j$, we construct a counterfactual trajectory by applying an intervention $\boldsymbol{\delta}_t$ to the system dynamics.

Let $\mathbf{x}_t^{(0)}$ denote the baseline trajectory generated by the fitted model, and $\mathbf{x}_t^{(\delta)}$ the trajectory under intervention. A one-time unit shock at horizon $h=1$ is defined as
\[
\boldsymbol{\delta}_t =
\begin{cases}
\delta\,\mathbf{e}_j, & t = t_0 + 1, \\
\mathbf{0}, & \text{otherwise},
\end{cases}
\]
where $\mathbf{e}_j$ is the $j$-th canonical basis vector and $\delta$ is scaled to the empirical standard deviation of variable $j$.

The counterfactual impulse response for variable $i$ at horizon $h$ is then defined as
\begin{equation}
\mathrm{IRF}_{i \leftarrow j}(t_0,h)
=
x^{(\delta)}_{i,t_0+h}
-
x^{(0)}_{i,t_0+h}.
\end{equation}

For system-level analysis, we summarize counterfactual deviation using the $L^2$ norm across variables:
\begin{equation}
\mathcal{R}(t_0+h)
=
\left\lVert
\mathbf{x}^{(\delta)}_{t_0+h}
-
\mathbf{x}^{(0)}_{t_0+h}
\right\rVert_2.
\end{equation}

Representative impulse responses and system-level counterfactual trajectories are selected for illustrative comparison across models in Section 5.

\section{Implementation Details for DCNAR}
\label{app:dcnar_impl}

This appendix documents implementation details of DCNAR to support reproducibility. All design choices reported here correspond to the experiments described in Section 5.

\subsection{Causal Discovery Stage}
\label{app:navar_impl}

\paragraph{NAVAR variant.}
The causal discovery stage is instantiated using a neural additive vector autoregression (NAVAR) model with additive, variable-specific neural components. Each target variable is modeled independently using a feedforward neural network with convolutional preprocessing over lagged inputs (MLP/Conv variant). No recurrent components are used.

\paragraph{Lag length.}
The maximum lag length is fixed to
\begin{equation}
L = 8,
\end{equation}
and identical lag structure is used for all variables and all countries.

\paragraph{Regularization.}
Sparsity in the learned causal graph is encouraged through $\ell_1$ regularization applied to the additive component functions. Specifically, regularization is applied to the contribution magnitudes of the neural functions $f_{ij\ell}$, encouraging many directed interactions to shrink toward zero. Weight decay is additionally applied to stabilize neural optimization.

\paragraph{Normalization.}
No normalization is applied to the input time series prior to NAVAR training. Likewise, no normalization is applied to the learned causal score matrix. All causal scores are therefore expressed in the native scale of the data and are treated as relative, not absolute, measures of influence.

\paragraph{Output.}
The output of the causal discovery stage is a directed causal score matrix
\[
\mathbf{S} \in \mathbb{R}^{N \times N},
\]
aggregated across lags. This matrix is treated as a structural hypothesis rather than as an inferential object. Additional processing of $\mathbf{S}$ (e.g., necessity filtering, stability screening) is described below.

\subsection{Construction of the Structural Prior via Edge Ablation}
\label{app:edge_ablation}

The purpose of refining the causal adjacency matrix is to obtain a sparse structural prior that prioritizes directed relationships which could materially affect model behavior. A detailed methodological treatment and evaluation of this approach is provided in \cite{Kuskova26}; here we document its role in the present experiments.

\paragraph{Edge ablation procedure.}
Starting from an initial directed adjacency matrix inferred during the causal discovery stage, we consider the effect of selectively removing individual directed edges. For each ordered pair $(j,i)$ such that the initial adjacency matrix indicates a potential connection, we construct an ablated variant of the network in which that single edge is removed while all other edges are retained.

For each ablated network, the dynamic model is re-estimated using the same specification and hyperparameters as the full model, differing only in the exclusion of the selected edge. This yields a family of models that are identical except for the presence or absence of a single directed interaction.

\paragraph{Forecast-based comparison.}
To assess the impact of edge removal, we compare out-of-sample predictive behavior of the full model and each ablated variant using a fixed forecasting protocol. Let $e_t^{(0)}$ denote the forecast error at time $t$ under the full model and $e_t^{(-ij)}$ the corresponding error under the model in which edge $(j,i)$ has been removed. The effect of ablation is summarized by the loss differential
\[
d_t^{(ij)} = L(e_t^{(-ij)}) - L(e_t^{(0)}),
\]
where $L(\cdot)$ is a fixed loss function.

Positive average loss differentials indicate that removal of the edge degrades predictive performance, while negligible or negative differentials suggest that the edge does not materially affect model behavior.

\paragraph{Statistical assessment.}
Statistical significance of loss differentials is evaluated using Diebold--Mariano tests \cite{Diebold16}, which account explicitly for serial dependence in forecast errors. These tests assess whether the predictive performance of the ablated model differs systematically from that of the full model over the evaluation period.

Edges whose removal leads to statistically significant degradation in predictive performance are retained in the structural prior. Edges whose removal has little or no effect are excluded. The resulting adjacency matrix is therefore weighted and sparse, reflecting predictive necessity rather than coefficient magnitude.

\paragraph{Dynamic coherence screening.}
As a supplementary diagnostic, we examine whether the resulting structural prior yields stable and interpretable time-varying dynamics. Specifically, we assess whether estimated causal influence trajectories exhibit smooth temporal evolution rather than erratic fluctuation. Structures that produce highly unstable or noisy dynamics are treated as unreliable and excluded from further analysis.

\paragraph{Role in the present paper.}
In the experiments reported in this paper, the structural prior supplied to the dynamic inference stage is the weighted adjacency matrix obtained via this edge ablation procedure. The procedure is used solely to select a plausible and empirically grounded structural prior; the substantive evaluation of edge necessity, stability, and coherence is addressed in detail in \cite{Kuskova26}.

\subsection{Dynamic Inference Stage}
\label{app:tvnar_impl}

\paragraph{tvNAR formulation.}
Dynamic causal inference is performed using a time-varying network autoregressive model \cite{Ding25} conditioned on a learned structural prior. The primary specification used in the main text is tvNAR(1), which takes the form
\[
\mathbf{x}_t = (\widehat{\mathbf{G}} + \mathbf{I})\,\boldsymbol{\Lambda}(t)\,\mathbf{x}_{t-1} + \boldsymbol{\varepsilon}_t,
\]
where $\widehat{\mathbf{G}}$ is the learned adjacency matrix and $\boldsymbol{\Lambda}(t)$ is a diagonal matrix of time-varying node influence parameters.

\paragraph{tvNAR($p$).} The implementation supports higher-order dynamics via tvNAR($p$),
\[
\mathbf{x}_t =
\sum_{\ell=1}^{p}
(\widehat{\mathbf{G}} + \mathbf{I})\,\boldsymbol{\Lambda}_\ell(t)\,\mathbf{x}_{t-\ell}
+ \boldsymbol{\varepsilon}_t,
\]
of which tvNAR(1) is a special case. In practice, tvNAR(1) is used in the main experiments, while tvNAR($p$) is explored in supplementary analyses.

\paragraph{Kernel smoothing.}
Time variation in $\boldsymbol{\Lambda}(t)$ is estimated using kernel-weighted local regression over a normalized time index $\tau \in (0,1)$. A Gaussian kernel is used:
\[
K(u) = \exp\!\left(-\tfrac{1}{2}u^2\right),
\]
with a fixed bandwidth parameter. Kernel weights are computed within each country series and then pooled across countries using panel-aware indexing.

\paragraph{Time indexing in panel data.}
Each country time series is mapped to a normalized time index
\[
\tau_{c,t} = \frac{t}{T_c},
\]
where $T_c$ is the length of the series for country $c$. Kernel smoothing is performed with respect to $\tau_{c,t}$, allowing time variation to be estimated consistently across panels of equal length. All panel splitting and indexing respect country boundaries; no temporal leakage occurs across units.

\subsection{Computational Considerations}
\label{app:compute}

\paragraph{Runtime.}
The computational cost of DCNAR is dominated by the causal discovery stage. NAVAR training scales approximately as
\[
\mathcal{O}(N^2 \cdot L \cdot T),
\]
where $N$ is the number of variables, $L$ the lag length, and $T$ the total number of time points across panels. In the experiments reported here, NAVAR training completes within two-three minutes on a single GPU. The dynamic inference stage scales linearly in $T$ for fixed $N$ and is computationally lightweight relative to NAVAR.

\paragraph{Memory usage.}
Memory requirements are modest. NAVAR requires storing lagged input tensors and per-variable neural models, while tvNAR requires only the learned adjacency matrix and time-varying coefficient paths. All experiments fit comfortably within standard GPU memory constraints.

\paragraph{Parallelization.}
The implementation supports parallelization across countries at both stages. NAVAR training is parallelized implicitly through batched optimization, while tvNAR estimation and counterfactual simulation are embarrassingly parallel across panels and forecast horizons. All reported experiments were run using standard multi-core CPU resources and a single GPU.

\paragraph{Reproducibility.}
All hyperparameters, data splits, and model configurations are fixed across experiments. No manual tuning is performed per country or per variable. The full experimental pipeline is deterministic up to stochastic neural optimization and Monte Carlo sampling used for uncertainty estimation.

\section{Extended Panel Analysis Setup}
\label{app:extended_panel}

This appendix documents the data construction and experimental protocol for the extended panel analysis used to assess the robustness of DCNAR to increased temporal support. Results corresponding to this setup are reported in Appendix~\ref{app:extended_results}.

\subsection{Extended Dataset Construction}
\label{app:extended_data}

The extended panel is derived from the Varieties of Democracy (V-Dem) country--year dataset and uses the same set of democracy components as the main analysis. Variable definitions, coding procedures, and substantive interpretation are identical to those described in Section 5.

Unlike the main panel, which prioritizes breadth across countries, the extended panel prioritizes temporal depth. Countries are retained only if they exhibit complete, uninterrupted coverage across all selected variables for a substantially longer time span. This requirement leads to a smaller but temporally richer sample. The resulting dataset consists of 89 countries, observed annually for $T = 75$ consecutive years, with no missing values across the selected democracy components.

This construction yields a strongly balanced panel with approximately twice the temporal length of the main analysis but fewer cross-sectional units. No additional filtering or transformation is applied beyond the criteria above.

\subsection{Motivation for Extended Panel Analysis}
\label{app:extended_motivation}

The extended panel serves as a robustness check for dynamic causal inference under substantially different data conditions. While the main analysis reflects the standard setting in empirical democracy research, where data are typically represented by many countries with relatively short time series, the extended panel approximates a complementary regime with longer temporal trajectories but reduced cross-sectional diversity.

Evaluating DCNAR under both configurations allows us to assess whether its inferred dynamic causal behavior depends critically on the short-panel setting or whether it reflects more persistent features of the data-generating process. In particular, the extended panel tests whether impulse responses and counterfactual trajectories remain stable when substantially more temporal information is available for each unit.

\subsection{Experimental Protocol}
\label{app:extended_protocol}

The experimental protocol for the extended panel mirrors that of the main analysis as closely as possible. All modeling choices, hyperparameters, and evaluation procedures are held fixed.

Specifically:
\begin{itemize}
\item The same causal discovery procedure is applied to the extended panel without retuning.
\item The same DCNAR dynamic inference configuration is used.
\item The same baseline models (Ridge VAR, TV-VAR, and LSTM with Monte Carlo dropout) are evaluated.
\item Training--evaluation splits are defined analogously, with each country series divided into a training segment and a held-out evaluation segment.
\end{itemize}

No model parameters are adjusted to account for the increased temporal length. This ensures that differences in behavior can be attributed to data characteristics rather than to changes in model specification.

\subsection{Results}
\label{app:extended_results}

This section reports results for the extended panel analysis based on the 75-year sample and interprets them in relation to the main results presented in Section~5 of the main paper. The purpose is to assess the robustness of DCNAR's dynamic causal behavior under substantially increased temporal support.

\begin{figure}[h]
  \centering
  \includegraphics[width=\linewidth]{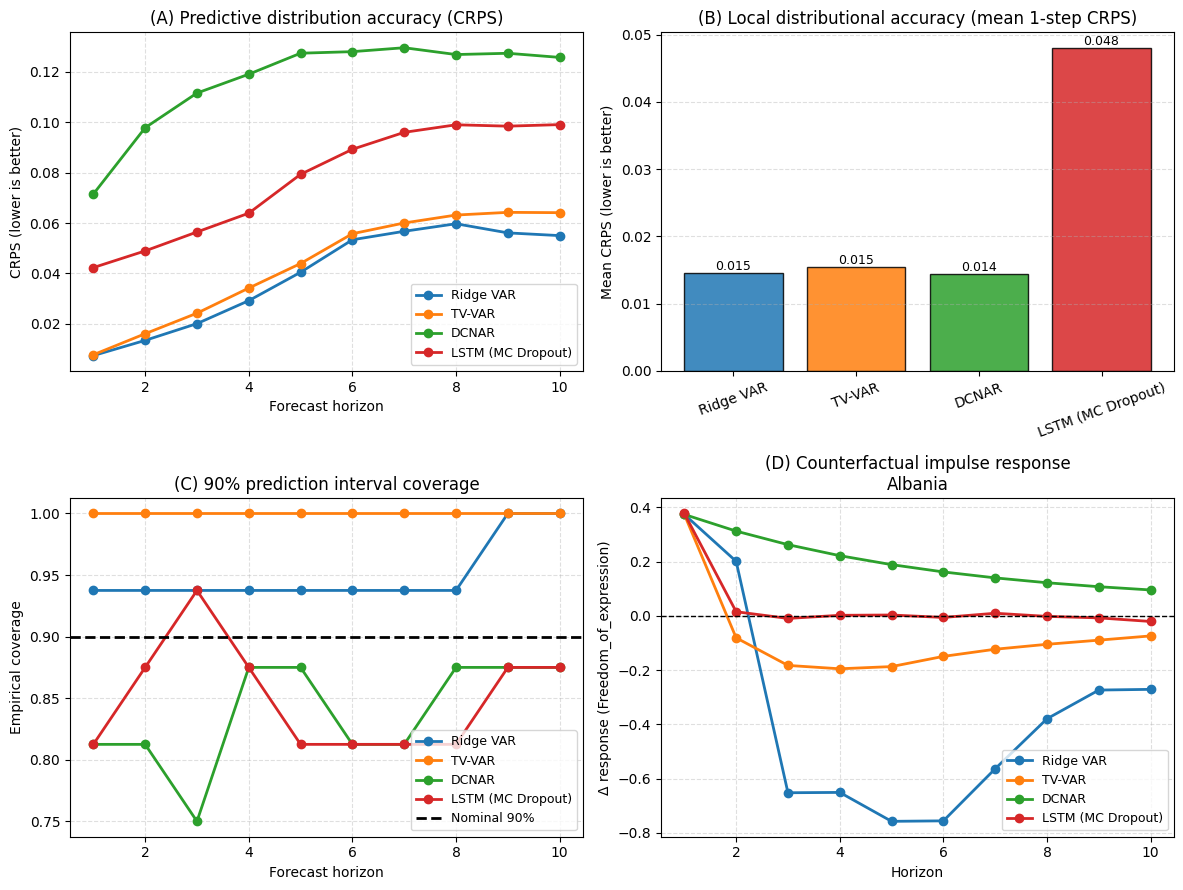}
  \caption{Comparison of DCNAR on the 89-country, 75-year panel, with Ridge VAR, TV-VAR, and LSTM (MC Dropout) across predictive (panel A, CRPS), distributional (panel B, local distributional accuracy - mean one-step-ahead CRPS), and causal diagnostics (panel C, nominal 90\% prediction intervals across horizons) on the extended-panel democracy dataset. Panel D shows representative counterfactual impulse response following a positive shock to freedom of expression (Albania).}
  \label{fig:supp_fig3}
  \Description{}
\end{figure}

\paragraph{Predictive and distributional performance.}
Panels (A)--(C) of Figure~\ref{fig:supp_fig3} summarize predictive distribution accuracy and calibration across models. As in the main analysis, DCNAR achieves predictive performance comparable to linear and time-varying VAR baselines across forecast horizons. Differences in CRPS and one-step distributional accuracy are modest, and no model consistently dominates across all horizons. Empirical coverage of nominal 90\% prediction intervals remains stable for DCNAR across horizons, while the LSTM baseline again exhibits undercoverage.

\begin{figure}[h]
  \centering
  \includegraphics[width=\linewidth]{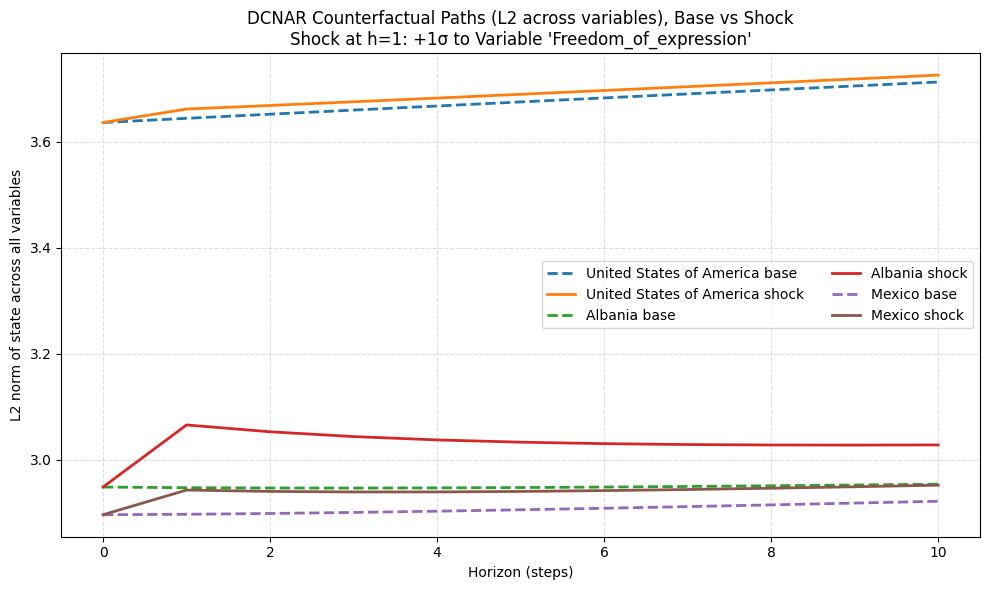}
  \caption{System-level counterfactual trajectories under DCNAR on the 89-country, 75-year panel, summarized by the $L^2$ normalization across all democracy components, for Albania, the United States, and Mexico. Solid lines denote trajectories under a localized positive shock to freedom of expression at horizon $h=1$; dashed lines denote corresponding baseline trajectories without intervention.}
  \label{fig:supp_fig4}
  \Description{}
\end{figure}

\paragraph{Impulse response and counterfactual behavior.}
Panel (D) of Figure~\ref{fig:supp_fig3} presents a representative counterfactual impulse response for Albania following a positive shock to freedom of expression; the response is smooth, monotonic, and gradually decaying, while Ridge VAR and TV-VAR again exhibit oscillations and sign reversals. Figure~\ref{fig:supp_fig4} extends the analysis to system-level counterfactual trajectories, summarized using the $L^2$ norm across all variables. Under DCNAR, counterfactual paths diverge smoothly from baseline and remain bounded over the forecast horizon, with cross-country magnitude differences reflecting known institutional heterogeneity.

The persistence of these qualitative differences in a substantially longer panel indicates that DCNAR's behavior is not an artifact of short time series or limited temporal resolution.

\paragraph{Implications for stability.}
Taken together, the extended panel results reinforce the central claim of the paper: DCNAR yields stable and interpretable dynamic causal behavior across data regimes with very different temporal characteristics. The consistency of impulse responses and counterfactual trajectories across the 35-year and 75-year panels suggests that DCNAR captures persistent features of the underlying data-generating process rather than overfitting to a particular sample window.

These findings support the interpretation of DCNAR as a robust methodological instrument for dynamic causal analysis in observational panel settings, rather than as a model whose conclusions are sensitive to specific data configurations.

\subsection{Scope and Interpretation}
\label{app:extended_scope}

The purpose of the extended panel analysis is not to improve predictive performance or to optimize model fit under favorable conditions. Rather, it is intended to assess the stability and robustness of dynamic causal behavior, in particular, impulse responses and counterfactual trajectories, when the amount of temporal information available per unit is substantially increased.

Results presented in Appendix~\ref{app:extended_results} should therefore be interpreted qualitatively, in comparison to the main analysis, with emphasis on consistency of dynamic patterns rather than on absolute performance metrics.

\section{Bandwidth Sensitivity Analysis}
\label{app:bandwidth}

\begin{figure}[h]
  \centering
  \includegraphics[width=0.9\linewidth]{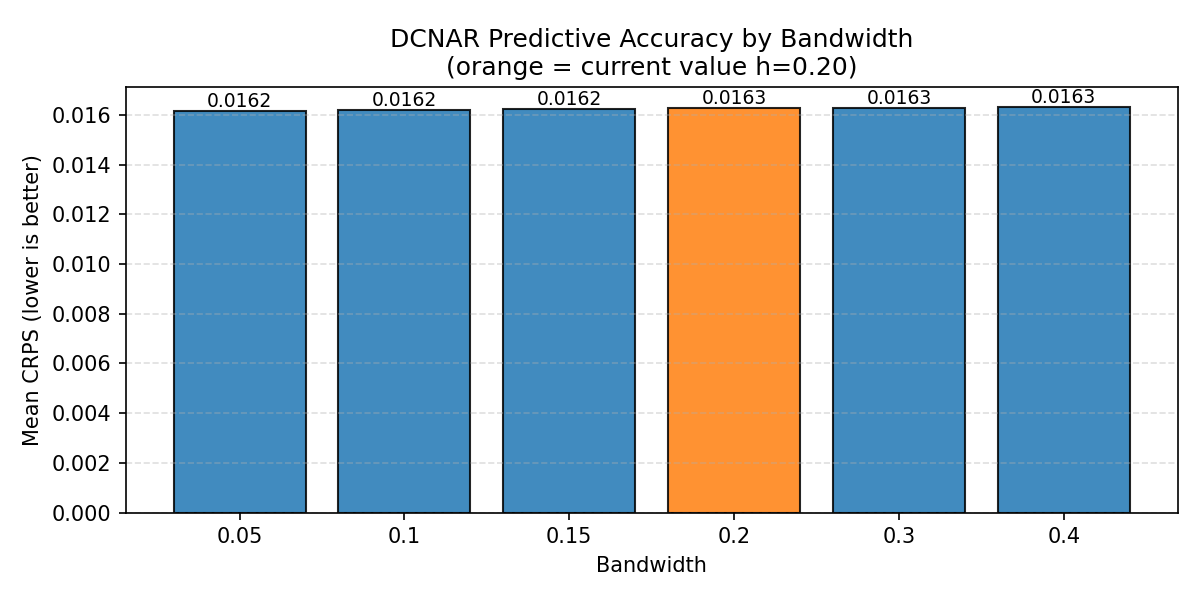}
  \caption{DCNAR predictive accuracy (mean CRPS) across bandwidth values. The
  bandwidth used in the main experiments ($h=0.20$) is highlighted. Predictive
  accuracy is essentially flat across the full range.}
  \label{fig:supp_bandwidth}
  \Description{}
\end{figure}

To verify that the smoothness of DCNAR's impulse responses reflects genuine
temporal regularization rather than an artifact of the kernel bandwidth, we
repeated the analysis across six bandwidth values spanning highly local to
near-global smoothing. The kernel is one-sided throughout: the estimate at time
$t$ uses only observations at times $s \le t$, so no future information
contributes to any coefficient estimate. Table~\ref{tab:bandwidth} reports mean
CRPS and impulse-response diagnostics; Figure~\ref{fig:supp_bandwidth} visualizes
the CRPS values.

\begin{table}[h]
\centering
\caption{Bandwidth sensitivity. Mean CRPS varies by less than $2\times10^{-4}$
across the full range, and impulse responses remain smooth, monotonic, and
sign-consistent (zero sign changes) for every country and bandwidth.}
\label{tab:bandwidth}
\small
\begin{tabular}{@{}lcccc@{}}
\toprule
Bandwidth $h$ & Mean CRPS & Sign consistent & Monotonic & Sign changes \\
\midrule
0.05 & 0.01616 & Yes & Yes & 0 \\
0.10 & 0.01622 & Yes & Yes & 0 \\
0.15 & 0.01624 & Yes & Yes & 0 \\
0.20 & 0.01626 & Yes & Yes & 0 \\
0.30 & 0.01629 & Yes & Yes & 0 \\
0.40 & 0.01630 & Yes & Yes & 0 \\
\bottomrule
\end{tabular}
\end{table}

The impulse-response diagnostics are identical across all bandwidths and all
three countries examined (Albania, the United States, and Mexico): smooth,
monotonic decay with a peak at horizon $h=1$ and zero sign changes. This confirms
that the qualitative dynamics reported in the main text are a feature of the
estimated system rather than a consequence of the smoothing choice.

\section{Architectural Ablation}
\label{app:ablation_supp}

\begin{figure}[h]
  \centering
  \includegraphics[width=\linewidth]{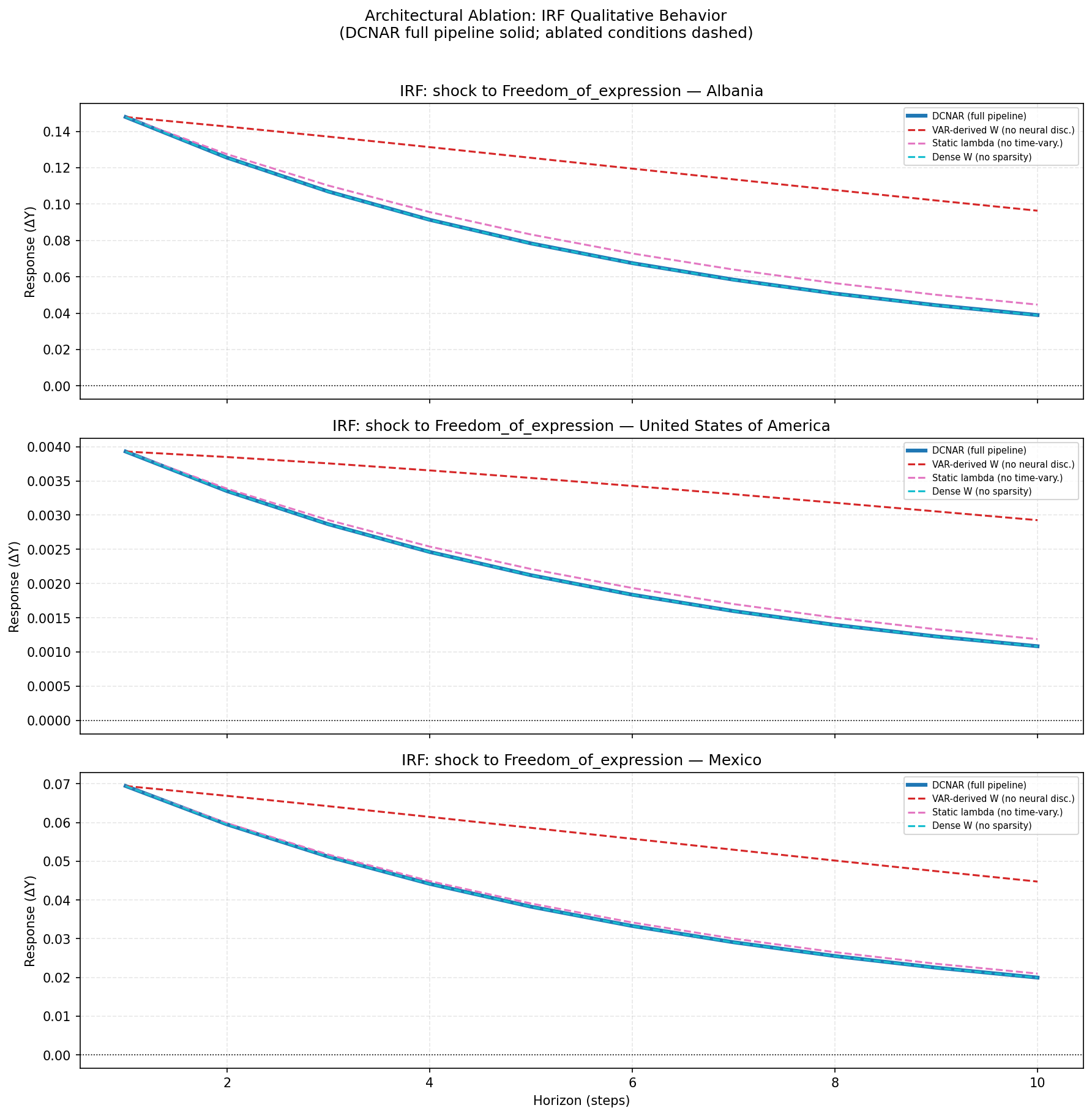}
  \caption{Impulse-response behavior under each ablation condition (DCNAR full
  pipeline solid; ablated conditions dashed) for shocks to freedom of expression
  in Albania, the United States, and Mexico. The VAR-derived structure
  systematically overestimates shock persistence (slower decay), while the other
  variants closely track the full pipeline.}
  \label{fig:supp_ablation}
  \Description{}
\end{figure}

This section provides the full architectural ablation summarized in
Table~1 (Section~5) of the main paper. Each DCNAR component is removed in turn
while all others are held fixed: (i) the neural causal-discovery module is
replaced by a Ridge VAR--derived adjacency at matched density; (ii) the
time-varying $\boldsymbol{\Lambda}(t)$ is replaced by its static, time-averaged
counterpart; and (iii) necessity-based sparsity is removed, yielding a dense
$\widehat{\mathbf{G}}$. Table~\ref{tab:ablation_full} reports predictive accuracy
and qualitative impulse-response diagnostics; Figure~\ref{fig:supp_ablation}
shows the corresponding per-country impulse responses.

\begin{table}[h]
\centering
\caption{Architectural ablation on the short panel. Replacing neural discovery
with a linear VAR--derived structure degrades CRPS by 14.5\%. All variants
preserve sign-consistent, monotonic impulse responses across the three countries
examined.}
\label{tab:ablation_full}
\small
\begin{tabular}{@{}lcccc@{}}
\toprule
Condition & Edges & $\boldsymbol{\Lambda}$ & CRPS & Sign / Mono. \\
\midrule
DCNAR (full pipeline) & 166 & time-varying & 0.01311 & 3/3 \,/\, 3/3 \\
VAR-derived $\widehat{\mathbf{G}}$ & 166 & time-varying & 0.01502 & 3/3 \,/\, 3/3 \\
Static $\boldsymbol{\Lambda}$ & 166 & static & 0.01306 & 3/3 \,/\, 3/3 \\
Dense $\widehat{\mathbf{G}}$ & 240 & time-varying & 0.01323 & 3/3 \,/\, 3/3 \\
\bottomrule
\end{tabular}
\end{table}

The neural discovery stage is the component with the clearest predictive
contribution: substituting a linear VAR--derived structure degrades CRPS by
14.5\% and produces impulse responses that systematically overestimate shock
persistence. Removing time variation leaves one-step CRPS essentially unchanged
(indeed marginally lower at 0.01306), which is expected in a highly persistent
panel where short-horizon prediction is insensitive to smooth coefficient drift;
the contribution of $\boldsymbol{\Lambda}(t)$ is visible instead in longer-horizon
causal behavior. Removing sparsity modestly worsens CRPS, consistent with
necessity-based filtering removing redundant propagation paths. Because
$\widehat{\mathbf{G}}$ and $\boldsymbol{\Lambda}(t)$ enter jointly through their
product, these results should be read as sensitivity analyses rather than fully
identified component effects.

\section{Time-Varying Coefficient Paths and Effective Networks}
\label{app:lambda}

This section provides the coefficient-path and effective-network visualizations
referenced in Section~6 of the main paper. Without any domain supervision, the
estimated node-influence paths $\lambda_i(t)$ recover historically documented
regime dynamics, and they do so more coherently than the corresponding TV-VAR
diagonal coefficients.

\begin{figure}[h]
  \centering
  \includegraphics[width=\linewidth]{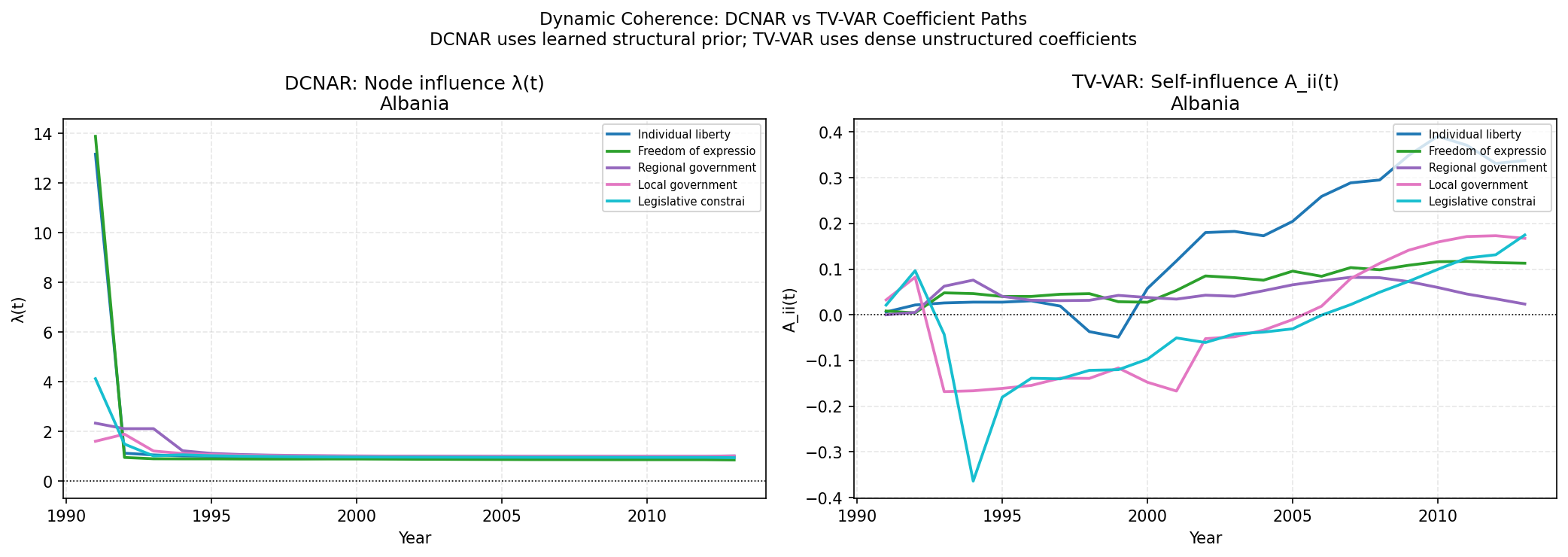}
  \caption{Dynamic coherence: DCNAR node-influence paths $\lambda(t)$ (left)
  versus TV-VAR self-influence coefficients $A_{ii}(t)$ (right) for Albania.
  DCNAR uses the learned structural prior; TV-VAR uses dense, unstructured
  coefficients. DCNAR paths stabilize after the early-1990s transition, whereas
  TV-VAR coefficients remain comparatively erratic.}
  \label{fig:supp_lambda}
  \Description{}
\end{figure}

For Albania, DCNAR influence parameters fall sharply over 1991--1993
(post-communist institutional collapse) before stabilizing near unity; for the
United States they remain near-constant (institutional stability); and for Mexico
they show a modest shift around 2000, coinciding with the end of single-party
rule. By contrast, the TV-VAR diagonal coefficient paths reach implausibly large
magnitudes for Mexico and Albania that are difficult to reconcile with domain
knowledge.

\begin{figure}[h]
  \centering
  \includegraphics[width=\linewidth]{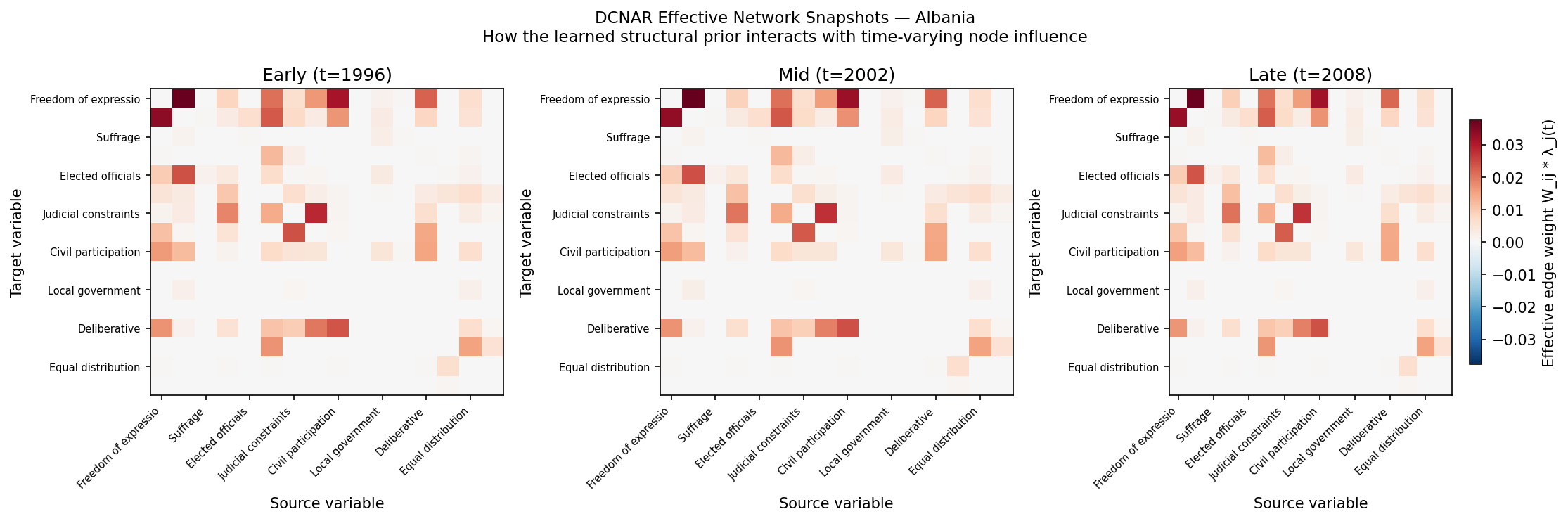}
  \caption{DCNAR effective network snapshots for Albania at three time points
  (1996, 2002, 2008), showing the effective edge weights $W_{ij}\,\lambda_j(t)$.
  The learned structure remains stable across the observation window while node
  influence evolves smoothly.}
  \label{fig:supp_network}
  \Description{}
\end{figure}

Effective-network snapshots (Figure~\ref{fig:supp_network}) confirm that the
learned structure remains stable across 1996, 2002, and 2008 while node influence
evolves smoothly. This kind of substantive, historically interpretable read-out
is inaccessible to static causal-discovery methods or black-box predictive models.

\end{document}